\documentclass[10pt,twocolumn,letterpaper]{article}

\usepackage{cvpr}
\usepackage{times}
\usepackage{epsfig}
\usepackage{graphicx}
\usepackage{amsmath}
\usepackage{amssymb}
\usepackage{hhline}
\usepackage{multirow}
\usepackage{array}

\usepackage[pagebackref=true,breaklinks=true,letterpaper=true,colorlinks,bookmarks=false]{hyperref}

\cvprfinalcopy 

\def\cvprPaperID{2187} 

\newcolumntype{C}[1]{>{\centering\let\newline\\\arraybackslash\hspace{0pt}}m{#1}}

\ifcvprfinal\fi
\begin{document}

\title{Annotating Object Instances with a Polygon-RNN}

\author{Llu\'is Castrej\'on \hspace{1cm} Kaustav Kundu \hspace{1cm} Raquel Urtasun \hspace{1cm} Sanja Fidler\\
Department of Computer Science\\
University of Toronto\\
{\tt\small \{castrejon, kkundu, urtasun, fidler\}@cs.toronto.edu}
}

\maketitle

\begin{abstract}
We propose an approach for semi-automatic annotation of object instances. 
While most current methods treat object segmentation as a pixel-labeling problem, we here cast it as a polygon prediction task, mimicking how most current datasets have been annotated. In particular, our approach takes as input an image crop and sequentially produces vertices of the polygon outlining the object. This allows a human annotator to interfere at any time and correct a vertex if needed, producing as accurate segmentation as desired by the annotator. We show that our approach speeds up the annotation process by a factor of 4.7 across all classes in Cityscapes, while achieving $78.4\%$ agreement in IoU with original ground-truth, matching the typical agreement between human annotators. 
For cars, our speed-up factor is 7.3 for an agreement of $82.2\%$. We further show generalization capabilities of our approach to unseen datasets.
\end{abstract}

\vspace{-3.4mm}
\section{Introduction}
\label{sec:intro}
\vspace{-1mm}

Semantic image segmentation has been receiving significant attention in the community~\cite{chen14semantic,LongCVPR2014}. With new benchmarks such as Cityscapes~\cite{cityscapes}, object instance segmentation is also gaining steam~\cite{IIS16,torr16,ZhangCVPR16,sharpmask,Uhrig16}.  Most of the recent approaches are based on neural networks, achieving impressive performance for these tasks~\cite{chen14semantic,LongCVPR2014,he15deepresidual,sharpmask}. Deep learning approaches are, however, data hungry and their performance is strongly correlated with the amount of available training data. This requires the community to annotate large-scale datasets which is both time consuming and expensive. Our goal in this paper is to make this process faster, while yielding ground-truth as precise  as the one available in the current datasets.

There have been several attempts at reducing the dependency on very detailed annotation such as object segmentation masks. In the weakly-supervised setting, approaches aim at learning segmentation models from weak annotation such as image tags or bounding boxes~\cite{Kuettel2012ECCV,JiaXu14,Jain16}.  In~\cite{Lin16}, the authors rely on scribbles, one on each object, while~\cite{ferrari16} requires only a single point on  the object. While these approaches hold promise, their performance is not yet competitive with fully supervised approaches. Other work exploits easier-to-obtain ground-truth such as bounding boxes, and produces (noisy) labeling inside each box with a GrabCut type of approach~\cite{Rother2004SIGGRAPH,ChenCVPR14}. It has been shown that such annotation can serve as useful auxilary data to train neural segmentation networks~\cite{ZhangICCV15,Uhrig16}. Yet, these segmentations cannot be used as official ground-truth for a benchmark due to its inherent imprecisions.

Most of the large-scale segmentation datasets have been collected by having annotators outline the objects with a polygon~\cite{pascal-voc-2010,mottaghirole,coco,cityscapes,ade20k}. Since typically objects are connected and without holes, polygons provide a way of annotating an object with a relatively small number of clicks, typically around 30 to 40 per object. In this paper, we propose an interactive segmentation method that produces highly accurate and structurally coherent object annotations, and reduces  annotation time by a factor of $4.7$.

\begin{figure}[t!]
\includegraphics[width=\linewidth,trim=0 94 60 0,clip]{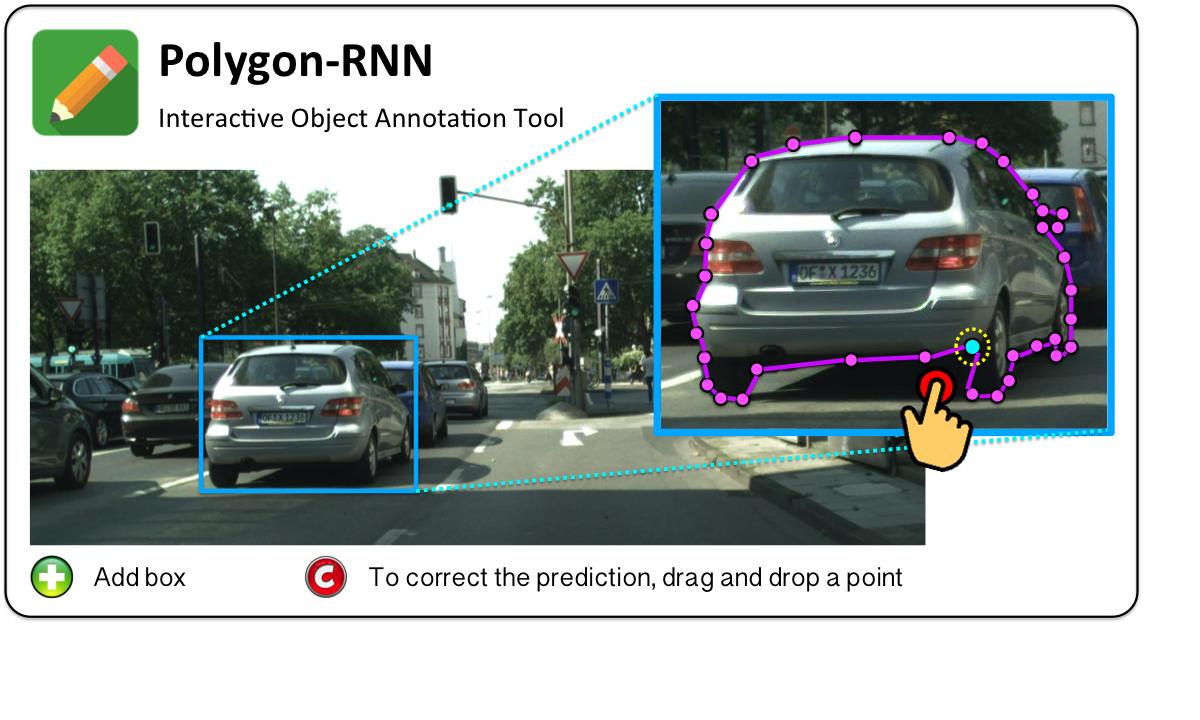} 
\caption{Given a bounding box, we automatically predict the polygon outlining the object instance inside the box, using our Polygon-RNN. Our method is designed to facilitation annotation, and easily incorporates user corrections of points to improve the overall object's polygon. Our method cuts down the number of required annotation clicks by a factor of $4.74$.}
\label{fig:intro}
\end{figure}

Given a ground-truth bounding box, our method generates a polygon outlining the object instance using a Recurrent Neural Network, which we call Polygon-RNN. Our approach takes as input an image crop and sequentially produces vertices of the polygon outlining the object. This allows a human annotator to interfere at any time and correct a vertex if needed, producing as accurate segmentations as desired by the annotator. We show that our annotation approach speeds up annotation process by factor of $4.7$, while achieving $78.4\%$ agreement with original ground-truth, matching the typical agreement of human annotators. We plan to release our code and create a web-annotation interface running our model at the backend. Please refer to our project page:{\small~\url{http://www.cs.toronto.edu/polyrnn}}. We hope this will cut down annotation time and cost of segmentation benchmarks in the future.

\vspace{-1mm}
\section{Related Work}
\label{sec:related}
\vspace{-1mm}

Our approach is related to work on semi-automatic image annotation and object instance segmentation.

{\bf Semi-automatic annotation}.  There has been significant effort at making pixel-level image labeling faster for the annotators.  In~\cite{Boykov2001ICCV},  the authors used scribbles as seeds  to model the appearance of foreground and background, and performed segmentation via graph-cuts by combining appearance cues and a smoothness term~\cite{Boykov2004PAMI}. ~\cite{Nagaraja} uses multiple scribbles on the object and background and exploits motion cues to annotate an object in a video. Scribbles were also recently used in~\cite{Lin16} to train CNNs for semantic image segmentation. GrabCut~\cite{Rother2004SIGGRAPH}  exploits annotations in the form of 2D bounding boxes, and performs per-pixel labeling with  foreground/background models  using EM. Building on top of this idea,~\cite{DeepCut} combined GrabCut with CNN to segment medical images. In~\cite{ChenCVPR14}, the authors exploited 3D bounding boxes and a point cloud to facilitate labeling. A different type of approach has been to exploit multiple bounding boxes and perform co-segmentation~\cite{Kuettel2012ECCV,Jain16}. 

Since most of these approaches define a graphical model at the pixel-level, with the smoothness term as the main relation among pixels, it is hard to incorporate shape priors. These are particularly important in ambiguous regions caused by shadows, image saturation or low-resolution of the object. Furthermore, nothing prevents these models to provide labelings with holes. If the method makes mistakes in outlining the object, the human annotator has a hard and tedious work to correct for such mistakes. Thus, these methods have mainly been used to produce additional, yet noisy training examples, but their output is typically not accurate enough to serve as official ground-truth of a benchmark. 

{\bf Annotation tools}.~\cite{Yamaguchi12} labeled clothing in images by performing annotation at the superpixel-level. This makes the labeling process more efficient, but inherently depends on the superpixel scale and thus typically merges small objects or parts. This issue was somewhat resolved in~\cite{Tuset15} by labeling videos at multiple superpixel scales. 


{\bf Object instance segmentation}. Our work is also related to object instance segmentation. Most of these approaches~\cite{IIS16,torr16,ZhangICCV15,ZhangCVPR16,deepmask,sharpmask} operate on the pixel-level, typically exploiting a CNN inside a box or a patch to perform the labeling. Work most related to ours is~\cite{ZhangCVPR12,Sun14} which aims to produce a polygon around an object. These approaches start by detecting edge fragments and find an optimal cycle that links the edges into a coherent region. In~\cite{Duan16}, the authors propose a method that produces superpixels in the from of small polygons which they combine into object regions with the aim to label aerial images. In our work, we predict the polygon around the object directly, using a carefully designed RNN.
\vspace{-2.4mm}
\section{Polygon-RNN}
\label{sec:method}
\vspace{-2mm}

Our goal is to create an efficient annotation tool for labeling object instances with polygons. As is typical in an annotation setting, we assume that the user provides the bounding box around the object. Given the image patch inside the box, our method predicts a (closed) polygon outlining the object using a Recurrent Neural Network. We allow the user to correct a predicted vertex of the polygon at any time step if needed, which we integrate in our prediction task.

We parametrize the polygon as a sequence of 2D vertices $(c_{t})_{t \in \mathbb{N}}$, $c \in \mathbb{R}^2$. We assume the polygon is closed, i.e., there is an edge between any two consecutive vertices, as well as the last and the first vertices. Note that a closed polygon is a cycle and thus has multiple equivalent parametrizations obtained by choosing any of the vertices as the beginning of the sequence, as well as selecting the orientation of the sequence. Here, we fix the polygon to always follow the clockwise orientation, but the starting point of the sequence can be any of the vertices. 

	

Our model is an RNN, that predicts a vertex at every time step. As input in each step of the RNN we use a CNN representation of the image crop, as well as the vertices  predicted one and two time steps ago, plus the first point. By explicitly providing information of the past two points we help the RNN to follow a particular orientation of the polygon. On the other hand, the first vertex helps the RNN to decide when to close (finish) the polygon. We train the RNN+CNN model end-to-end. This essentially helps the CNN to be fine-tuned to object boundaries, while the RNN learns to follow these boundaries and exploits its recurrent nature to also encode priors on object shapes. Our model thus returns a structurally coherent representation of the object. We name our model Polygon-RNN.



\begin{figure*}[t]
\vspace{-2mm}
	\centering
	\includegraphics[width=\textwidth,trim=0 65 0 15,clip]{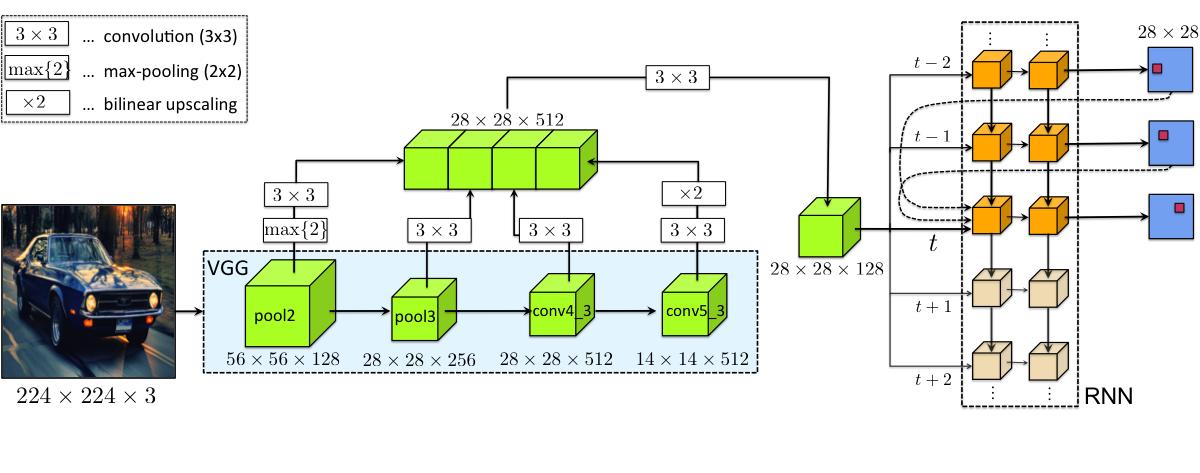}
	\caption{{\bf Our Polygon-RNN model}. At each time step of the RNN-decoder (right), we feed in an image representation using a modified VGG architecture. Our RNN is a two-layer convolutional LSTM with skip-connection from one and two time steps ago.  At the output at each time step, we predict the spatial location of the new vertex of the polygon.}
	\label{fig:model}
	\vspace{-0.2cm}
\end{figure*}

Figure~\ref{fig:model} shows the overview of the model. We next describe each component of the model in more detail.

\vspace{-1.2mm}
\subsection{Model Architecture}
\vspace{-1.8mm}

We start by providing details on the image representation via a CNN, and then explain the design of the RNN.

\vspace{-3mm}
\subsubsection{Image Representation via a CNN with Skip Connections}
\label{sec:cnn}
\vspace{-2mm}

We adopt the VGG-16 architecture~\cite{vggcnn} and modify it for the purpose of our task. We first remove the fully connected layers as well as the last max-pooling layer, \textit{pool5}. The output of this modified network has a downsampling factor of 16. We then add additional convolutional layers with skip-connections that fuse information from the previous layers and upscale the output by factor of 2 (downsampling factor of 8 wrt to the original size of the image crop, which is always scaled to $224\times 224$). This allows the CNN to extract features that contain both low-level information about the edges and corners, as well as semantic information about the object. The latter helps the model to ``see'' the object, while the former helps it to follow the object's boundaries.

We employ a similar architecture for the skip-connections as the one in~\cite{sharpmask}. The design guideline is to first process the features in the skip-layers using another convolutional layer, then concatenate all  outputs, and finally process this concatenated feature using another convolutional layer. We employ convolutional filters with a kernel size of $3 \times 3$, followed by a ReLU non-linearity. Concatenation layers join the channels of different outputs into a single tensor. Since we use features from multiple skip-layers which have different spatial dimensions, we employ bilinear upsampling or max-pooling in order to get outputs that all have the same spatial resolution. We refer the reader to Fig.~\ref{fig:model} for a visualization and further details about the architecture (the CNN is highlighted in green).

\vspace{-3mm}
\subsubsection{RNN for Vertex Prediction}
\label{sec:rnn}
\vspace{-2.4mm}

An RNN is a powerful representation of time-series data, as it carries more complex information about the history by employing linear and non-linear functions. In our case, we hope the RNN to capture the shape of the object and thus make coherent predictions even in ambiguous cases such as for example shadows and saturation. 

In particular, we employ a Convolutional LSTM~\cite{convlstm} in our model, and use it as a decoder. ConvLSTMs operate in 2D, which allows us to preserve the spatial information received from the CNN. Furthermore, a ConvLSTM employs convolutions, thus greatly reducing the number of parameters to be learned compared to using a fully-connected RNN. In its simplest form, a ConvLSTM (single layer) computes the hidden state $\mathbf{h_t}$ given the input $\mathbf{x_t}$ according to the following equations:
\begin{align}
\label{lstm}
	\left(
	\begin{array}{c}
		\mathbf{i_t} \\ \mathbf{f_t} \\ \mathbf{o_t} \\ \mathbf{g_t} \\
	\end{array}
	\right)
	&=
	\mathbf{W_h} * \mathbf{h_{t-1}} + \mathbf{W_x} * \mathbf{x_t} + \mathbf{b} \\
	\mathbf{c_t} &= \sigma(\mathbf{f_t}) \odot \mathbf{c_{t-1}} + \sigma(\mathbf{i_t}) \odot \tanh(\mathbf{g_t}) \nonumber\\
	\mathbf{h_t} &= \sigma(\mathbf{o_t}) \odot \tanh(\mathbf{c_t}) \nonumber
\end{align}
Here $i$, $f$, $o$ denote the input, forget, and output gate, $h$ is the hidden state and $c$ is the cell state. $\sigma$ denotes the sigmoid function, $\odot$ indicates an element-wise product and $*$ a convolution. $W_h$ denotes the hidden-to-state convolution kernel and $W_x$  the input-to-state convolution kernel.


In particular, we model the polygon with a two-layer ConvLSTM with kernel size of $3 \times 3$ and 16 channels, which outputs a vertex at each time step. We formulate the vertex prediction as a classification task. Specifically, we represent our output at time step $t$ as one-hot encoding of a $D \times D + 1$ grid, where the $D \times D$ dimensions represent the possible 2D positions of the vertex, and the last dimension corresponds to the end-of-sequence token (i.e., polygon is closed). The position of the vertices are thus quantized to the resolution of the output grid. Let $y_t$ denote the one-hot encoding of a vertex, output at time step $t$.

Our ConvLSTM gets as input a tensor ${\bf x}_t$ at time step $t$, that concatenates multiple features: the CNN feature representation of the image, $y_{t-1}$ and $y_{t-2}$, i.e., a one-hot encoding of the previous predicted vertex and the vertex predicted from two time steps ago, as well as the one-hot encoding of the first predicted vertex $y_1$.

Given two consecutive vertices,  the next vertex on the polygon is uniquely defined. However, this is not the case for the first vertex, since any vertex of the polygon can serve as a starting point (polygon is a cycle). We thus treat the starting point as special, and predict it in the following way. 
We reuse the same architecture of the CNN as in Sec.~\ref{sec:cnn}, but add two layers, each of dimension $D\times D$. One branch predicts object boundaries while the other takes as input the output of the boundary-predicting layer as well as the image features and predicts the vertices of the polygon. We treat both, the boundary and vertices as a binary classification problem in each cell in the output grid.



\vspace{-0.7mm}
\subsection{Training}
\label{sec:training}
\vspace{-1.mm}

To train our model we use cross-entropy at each time step of the RNN. In order to not over-penalize the incorrect predictions that are close to the ground-truth vertex, we smooth our target distribution at each time step. We assign non-zero probability mass to those locations that are within a distance of 2 in our $D \times D$ output grid. 

We follow the typical training regime where we make predictions at each time step but feed in ground-truth vertex information to the next. 
We train our model using the Adam optimizer~\cite{adamopt} with a batch size $b = 8$ and an initial learning rate of $\lambda = 1e-4$. We decay the learning rate after 10 epochs by a factor of 10 and use the default values of $\beta_1 = 0.9$ and $\beta_2 = 0.999$. 

For the task of first vertex prediction, we train another CNN using a multi-task loss. In particular, we use the logistic loss for every location in the grid. As ground-truth for the object boundaries, we draw the edges of the ground-truth polygon, and use the vertices of the polygon as ground-truth for the vertex layer. Our full model takes approximately a day to train on a Nvidia Titan-X GPU.  



\subsection{Inference and Annotators in the Loop}
\vspace{-1.8mm}
 
Inference in our model is done by taking the vertex with the highest log-prob at each time step of the RNN. This allows for a simple annotation interface: the annotator can correct the prediction at any time step, and we feed in the corrected vertex to the next time-step of the RNN (instead of the prediction). This puts the model back "on the right track". Typical inference time is 250 ms per object.  
 
\subsection{Implementation details}
\vspace{-1.8mm}
\label{sec:implementation}

We predict the polygon at resolution $D\times D$. In our experiments we used $D=28$, corresponding to an 8x downsampling factor with the input resolution and matching the resolution of the ConvLSTM. We perform polygon simplification with zero error in the quantized grid to eliminate vertices that lie on a line and to remove multiple vertices that would fall in the same grid position as a result of the quantization process.



We perform three different types of data augmentation: (1) we randomly flip the image crop and the corresponding polygon annotation, (2) we randomly select the amount of context expansion (enlarging the box) between 10\% and 20\% of the original bounding box and (3) we randomly select the starting vertex of our polygon annotation.


\vspace{-1.5mm}
\section{Results}
\label{sec:results}
\vspace{-1mm}

\begin{figure}[t!]
\vspace{-1mm}
	\centering
	\includegraphics[scale=0.292]{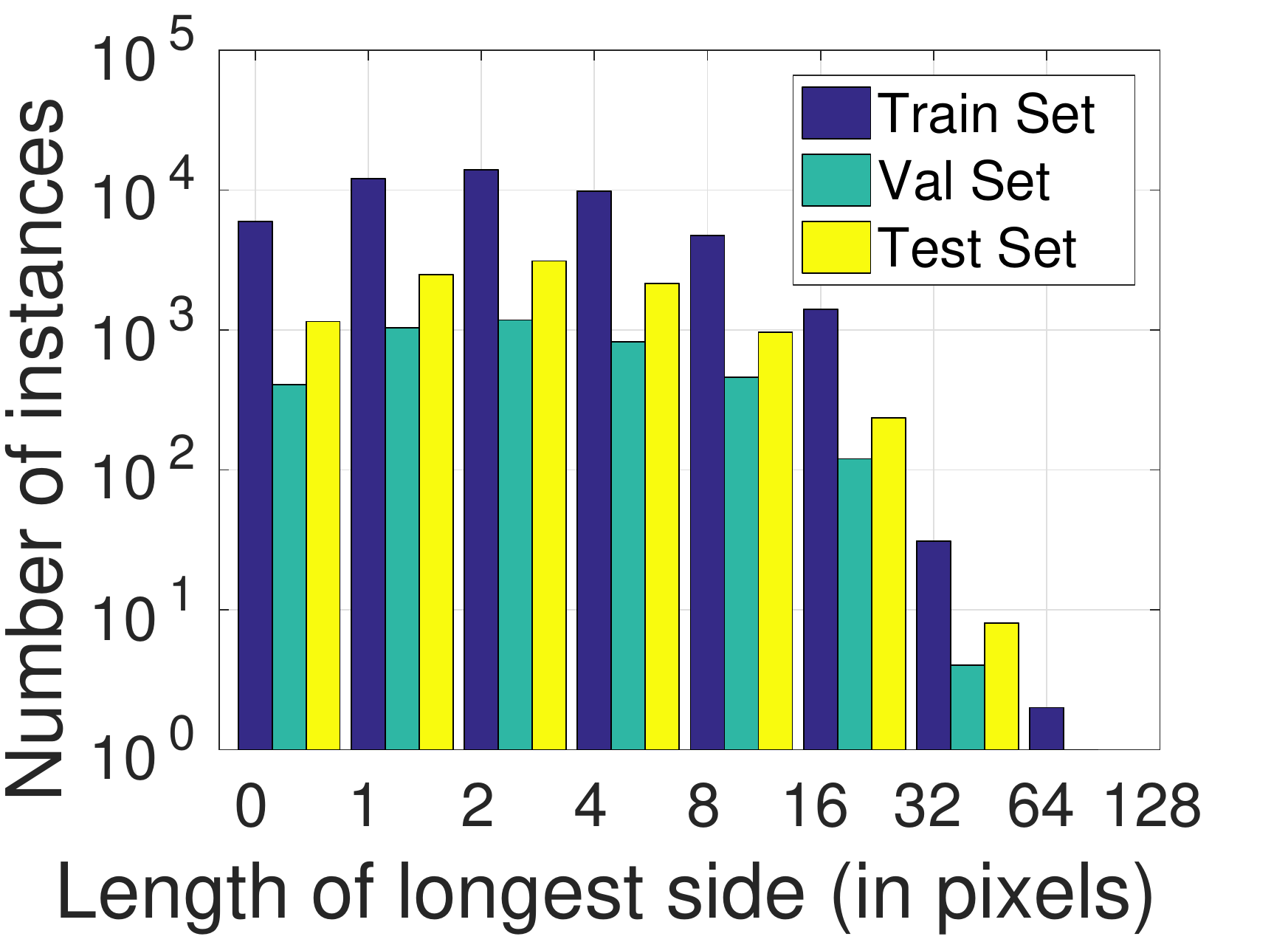}
	\caption{\textbf{Distribution of instances across different sizes:} The longest side on the X axis are multiples of 28 pixels}
	\label{fig:instance_size}
\end{figure}

\begin{figure}
\vspace{-2.5mm}
	\centering
	\includegraphics[scale=0.358]{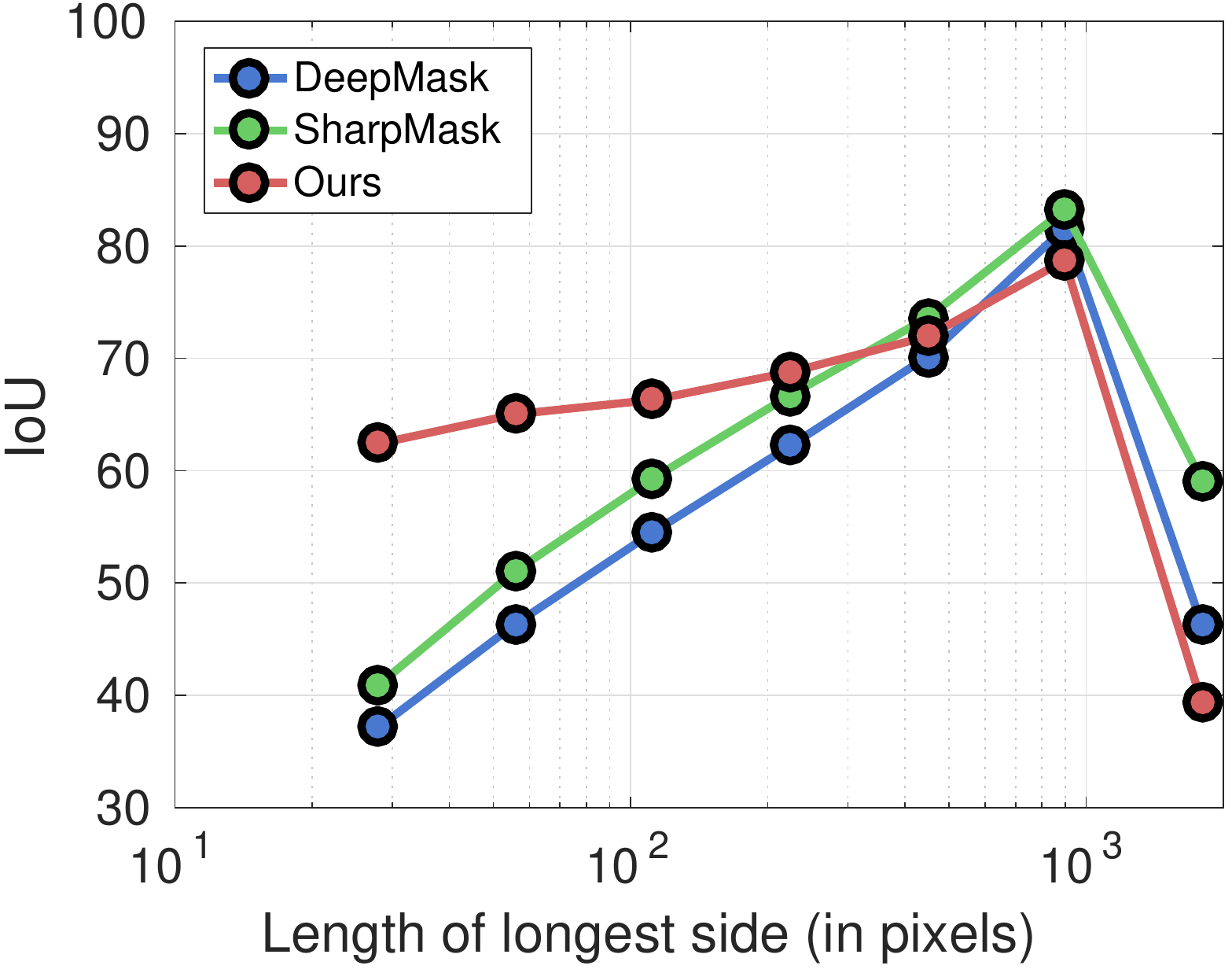}
	\caption{\textbf{IoU vs size of instance} comparing different approaches. Here, ours is run in prediction mode. }
	\label{fig:iou_size}
	\vspace{-2mm}
\end{figure}

We evaluate our approach for the task of object instance annotation on the Cityscapes dataset~\cite{cityscapes}, and provide additional results on KITTI~\cite{kitti}. Note that in all our experiments we assume to be given a ground-truth box around the object. Our goal then is to provide a polygon outlining this object  as accurately as possible and with minimal number of clicks required from the annotator. We report our performance with the standard IOU measure, as well as the number of vertex corrections of the predicted polygon. A box around the object in principle requires two additional clicks. However, boxes are typically much easier and cheaper to obtain using crowd-sourcing services such as AMT, while for most major segmentation benchmarks, polygons have been collected with high quality (in-house) annotators. 

\begin{table*}[t!]
\vspace{-2mm}
	\centering
	\begin{minipage}{0.49\linewidth}
	\begin{small}
	\addtolength{\tabcolsep}{-4.5pt}
	\begin{tabular}{|c||c||c|c|c|c|c|c|c|c|}
		\hline
		Split & \# Img. & Person & Rider & Car & Truck & Bus & Train & Mbike & Bike \\
		\hline
		Train & 2711 & 16452 & 1575 & 24982 & 455 & 352 & 136 & 657 & 3400 \\ 
		Val. & 264 & 1462 & 180 & 1962 & 27 & 27 & 32 & 78 & 258 \\ 
		Test & 500 & 3395 & 544 & 4658 & 93 & 98 & 23 & 149 & 1167 \\ 
		\hline
	\end{tabular}
	\end{small}
	\vspace{-0.5mm}
	\caption{Number of object instances per class in Cityscapes.}
	\label{table:data-stats}
\end{minipage}
\hspace{2mm}
	\begin{minipage}{0.485\linewidth}
	\vspace{-0.0mm}
	\begin{small}
	\addtolength{\tabcolsep}{-4.6pt}
	\begin{tabular}{|c||c|c|c|c|c|c|c|c||c|}
		\hline
		Mode & Car & Truck & Train & Bike & Prsn. & Rider & Mbike & Bus &Avg. \\
		\hline
		Comp-wise & 24.3 & 27.2 & 23.6 & 24.2 & 27.9 & 31.6 & 29.2 & 26.1 & 26.8 \\
		Inst-wise & 31.7 & 41.7 & 66.6 & 40.0 & 35.0 & 44.7 & 45.7 & 50.8 & 44.5 \\ 
		\hline
	\end{tabular}
	\end{small}
	\caption{Average number of vertices in polygon annotations for different object classes in Cityscapes.}
	\label{table:cityscapes-length}
	\end{minipage}
		\vspace{-2mm}

\end{table*}


\subsection{Cityscapes Dataset}
\vspace{-0.84mm}
We evaluate our approach on the Cityscapes instance segmentation dataset~\cite{cityscapes}. This dataset has images taken from  27 cities in Germany and neighboring countries. It contains 2975 training, 500 validation and 1525 test images. Since we do not have ground truth instances on the test set, we use an alternative split, where the 500 original validation images form our test set. We then split the original training set and select the images from two cities (Weimar and Zurich) as our validation, while the remaining cities become our training set. 
The dataset has annotations for eight object categories: \emph{person}, \emph{rider}, \emph{car}, \emph{truck}, \emph{bus}, \emph{train}, \emph{motorcycle} and \emph{bicycle}. The number of instances for each of these classes in our split is shown in Table~\ref{table:data-stats}. The Cityscapes dataset has instances with a large variation in their sizes. We show the distribution of instances for different lengths of the longest side of the box, in Fig.~\ref{fig:instance_size}. We observe a large variance, from 28 pixels to 1792 pixels.

\begin{table*}[t!]
	\centering
	\begin{tabular}{|c||c|c|c|c|c|c|c|c||c|}
		\hline
		Model & Bicycle & Bus & Person & Train & Truck & Motorcycle & Car & Rider & Mean \\
		\hhline{|=#=|=|=|=|=|=|=|=#=|}
		Square Box & 35.41 & 53.44 & 26.36 & 39.34  & 54.75 & 39.47 & 46.04 & 26.09 & 40.11 \\
		Dilation10 & 46.80 & 48.35 & 49.37 & 44.18 & 35.71 & 26.97 & 61.49 & 38.21 & 43.89 \\
		DeepMask~\cite{deepmask} & 47.19 & 69.82 & 47.93 & 62.20 & 63.15 & 47.47 & 61.64 & 52.20 & 56.45 \\
		SharpMask~\cite{deepmask} & 52.08 & \textbf{73.02} & 53.63 & {\bf 64.06} & 65.49 & 51.92 & 65.17 & 56.32 & 60.21 \\
		\hline
		Ours & {\bf 52.13} & 69.53 & {\bf 63.94} & 53.74 & {\bf 68.03} & {\bf 52.07} & \textbf{71.17} & {\bf 60.58} & {\bf 61.40} \\
		\hline
	\end{tabular}
	\vspace{1mm}
	\caption{\textbf{Performance} (IoU in \%) on all the Cityscapes classes {\bf without the annotator in the loop}.}
	\label{table:cityscapes-iou}
	\vspace{-2.4mm}
\end{table*}

Cityscapes provides instance segmentation ground truth both in terms of a pixel labeling as well as in terms of polygons. In the former, each pixel can correspond to at most one instance, thus representing the visible portion of the object. However, Cityscapes' polygons typically also capture some occluded parts of an instance, since the annotation tool performed depth ordering of objects to effectively remove the occluded portions~\cite{cityscapes}. We process the polygons to recreate the layering effect and obtain polygons representing only the visible portions of each object. The average number of vertices from the resulting polygons are shown in Table~\ref{table:cityscapes-length}. Since objects can be broken into multiple components due to occlusion, component-wise statistics  treats each component as a single example, while instance-wise statistics treats the entire instance as an example. Based on this statistics, we choose a hard limit of 70 time steps for our RNN, taking also GPU memory requirements into account.

\vspace{-4.2mm}
\paragraph{Evaluation Metrics:} We measure two aspects of our predicted annotations. For evaluating their quality, we use the intersection over union (IoU) metric, computed on a per-instance basis, and averaging across all instances. This is a strict measure since the small objects are penalized the same as the large instances. For evaluating the amount of human action required to correct our annotations, we simulate an annotator that corrects a point each time the predicted vertex deviates from the GT vertex more than a threshold. We then report the number of corrections (measured as clicks).

\vspace{-1.2mm}
\subsection{Prediction Mode}
\vspace{-1.2mm}

We first sanity check the performance of our model without any interaction from the annotator, i.e., we predict the full polygon automatically. We will refer to this setting as the \emph{prediction mode}.

\vspace{-3.6mm}
\paragraph{Baselines:} We use the recently proposed DeepMask~\cite{deepmask} and SharpMask~\cite{sharpmask} as state-of-the-art baselines. Given an input image patch, DeepMask uses a CNN to output a pixel labeling of an object, and does so agnostic to the class. Sharpmask extends Deepmask by clever upsampling of the output to obtain the labeling at a much higher resolution (160 vs 56). Note that in their original approach, ~\cite{deepmask, sharpmask} exhaustively sample patches at different scales over the entire image. Here, we use ground-truth boxes when reporting performance for their approach. Further, DeepMask and SharpMask use a 50 layer ResNet~\cite{he15deepresidual} architecture, which has been trained on the COCO~\cite{coco} dataset. We fine-tune this network on our Cityscapes split in two steps. In the first step, we fine-tune the feed-forward ResNet architecture for 150 epochs, followed by fine-tuning the weights for the Sharpmask's upsampling layers, for 70 epochs. This two step process is in the same spirit as that suggested in the paper.
Note that while these two approaches perform well in labeling the pixels, their output cannot easily be corrected by an annotator in cases when mistakes occur. This is in contrast to our approach, which efficiently integrates a human in the loop in order to get high quality annotations. 

We use two additional baselines, SquareBox and Dilation10. 
SquareBox is a simple baseline where the full box is labeled as the object. Instead of taking the tight-fit box, we reduce the dimensions of the box, keeping the same aspect ratio. Based on the validation set, we get the best results by choosing 80\% of the original box. If an instance has multiple components, we fit a box for each individual component as opposed to using the full box. This baseline mimics the scenario, in which the object is modeled simply as a box rather than a polygon. For the Dilation10 baseline, we use the segmentation results from~\cite{dilation}, which was trained on the Cityscapes segmentation dataset. For each bounding box, we consider the pixels belonging to the respective object category as the instance mask.



\vspace{-3mm}
\paragraph{Quantitative Results: } We report the IoU metric  in Table~\ref{table:cityscapes-iou}. We outperform the baselines in 6 out of 8 categories, as well as in the average across all classes. We perform particularly well in \textit{car}, \textit{person}, and \textit{rider}, outperforming Sharpmask by 12\%, 7\%, and 6\%, respectively. This is particularly impressive since Sharpmask uses a more powerful ResNet architecture (we use VGG). 


\vspace{-3.6mm}
\paragraph{Effect of object size:} In Fig.~\ref{fig:iou_size}, we see how our model performs w.r.t baselines on different instance sizes. For small instances our model performs significantly better than the baselines. For larger objects, the baselines have an advantage due to larger output resolution. This effect is most notable for classes such as bus and train, in which our model obtains lower IOU compared to the baselines.

\subsection{Annotator in the loop}
\vspace{-1.0mm}

The main advantage of our model is that it allows a human annotator to easily interfere if a mistake occurs. In particular, at each RNN time step, the annotator has the possibility to correct a misplaced  vertex. The correction is fed to the model at the next time step replacing the model's prediction, effectively helping the model to get back to the right track. Our goal is to obtain high quality annotations while minimizing annotation time.

We analyze how many clicks are needed to obtain different levels of segmentation accuracy. We perform such analysis by simulating an annotator: we correct a prediction if it deviates from the ground truth vertex by a certain distance. Distances are computed at the model output resolution using the \textit{chessboard} metric. In our experiments we compare the corrected predictions using  distance thresholds $T \in [1, 2, 3, 4]$.
In Table~\ref{table:annotation} we show the resulting IoU given different thresholds on the distance. We can observe a trade-off between the number of corrections and these metrics. 

\begin{center}
\begin{figure*}
\vspace{-3mm}
\centering
    \begin{tabular}{c c c c} 
        \hspace{7mm}{\small\sf\bf Bicycle} & \hspace{5mm}{\small\sf\bf Bus} & \hspace{5mm}{\small\sf\bf Person} & \hspace{1mm}{\small\sf\bf Train} \\[-0.7mm] \hspace{-0.4cm}
        \includegraphics[height=3.3cm,width=0.25\linewidth,trim=0 32 0 0,clip]{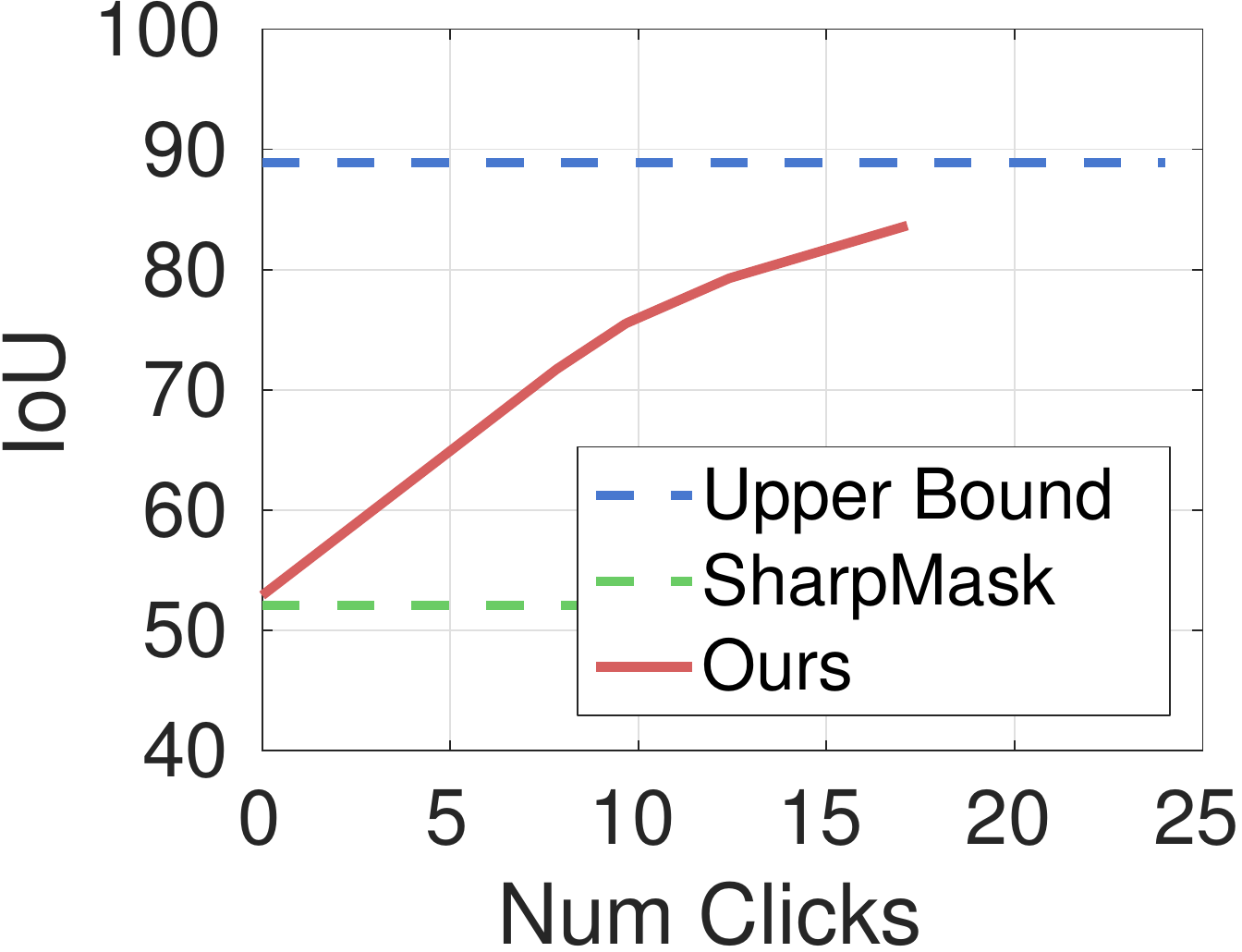} \hspace{-0.5cm} & 
        \includegraphics[height=3.3cm,width=0.25\linewidth,trim=32 32 0 0,clip]{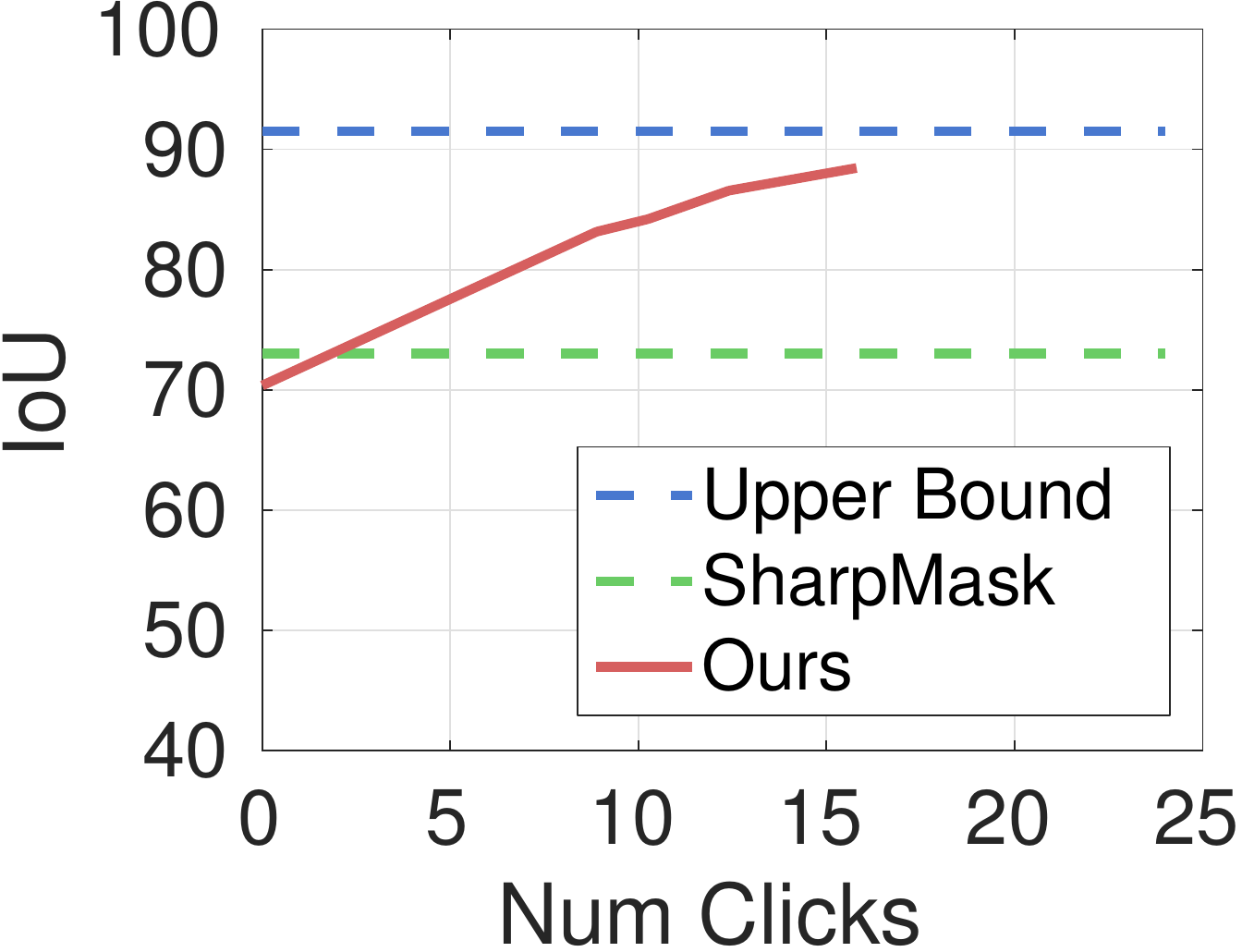} \hspace{-0.5cm} & 
        \includegraphics[height=3.3cm,width=0.25\linewidth,trim=32 32 0 0,clip]{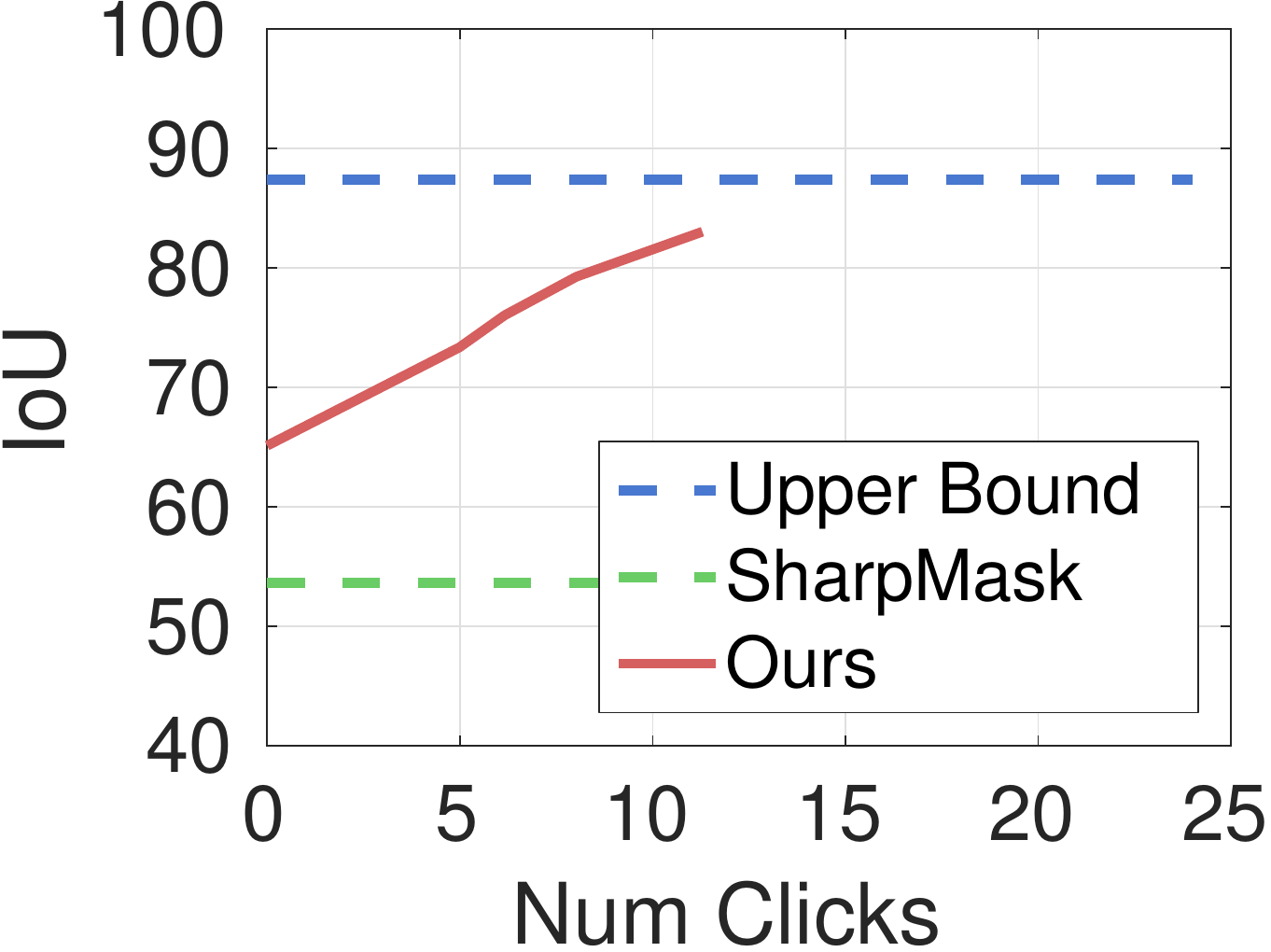} \hspace{-0.5cm} & 
        \includegraphics[height=3.3cm,width=0.25\linewidth,trim=32 32 0 0,clip]{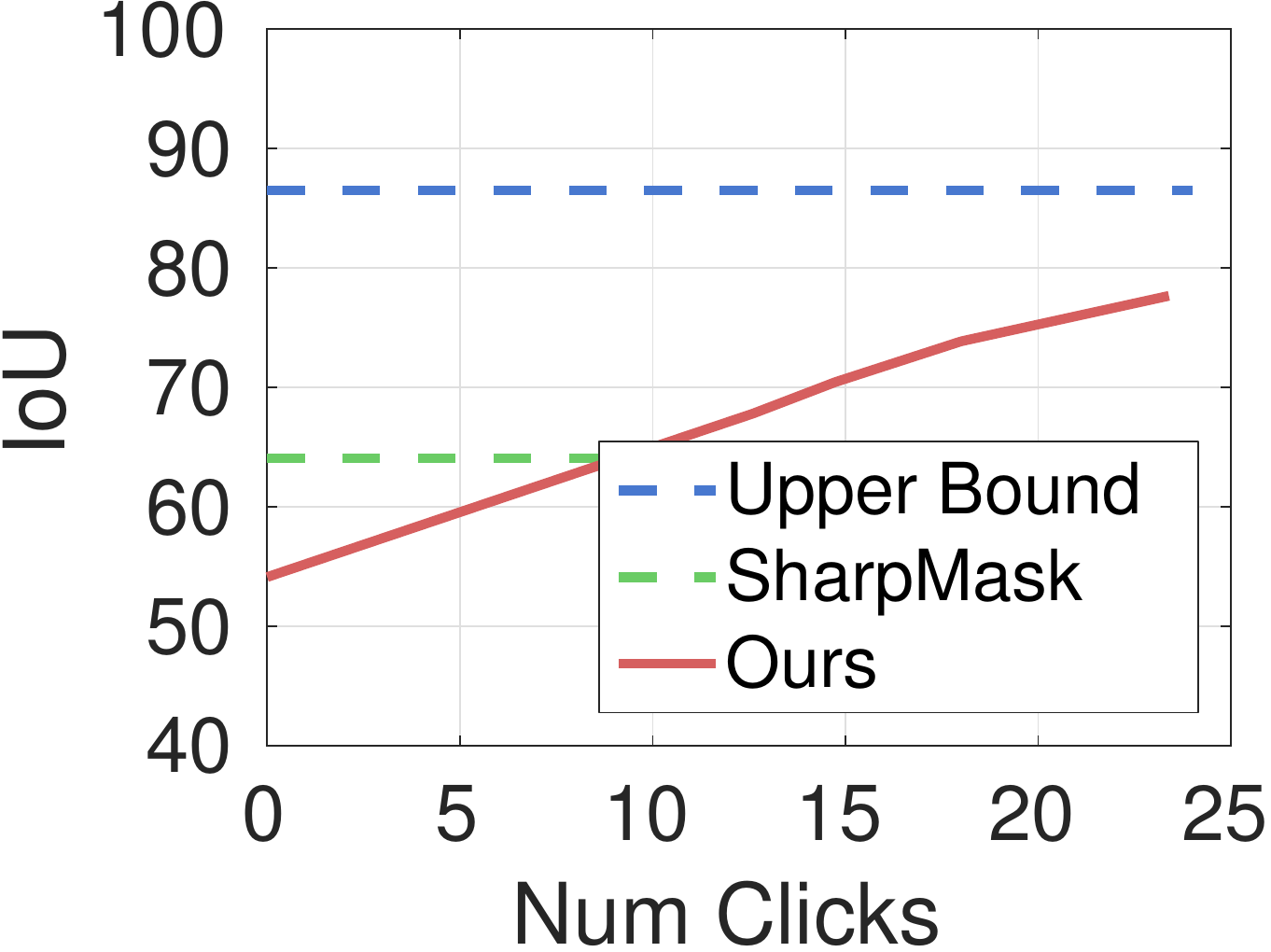} \vspace{-0.45cm} \\
        \\[0.5mm] 
        \hspace{7mm}{\small\sf\bf Truck} & \hspace{5mm}{\small\sf\bf Motorcycle} & \hspace{5mm} {\small\sf\bf Car} & \hspace{1mm}{\small\sf\bf Rider} \vspace{-0.48cm} \\
        \\ \hspace{-0.4cm}
        \includegraphics[height=3.6cm,width=0.25\linewidth]{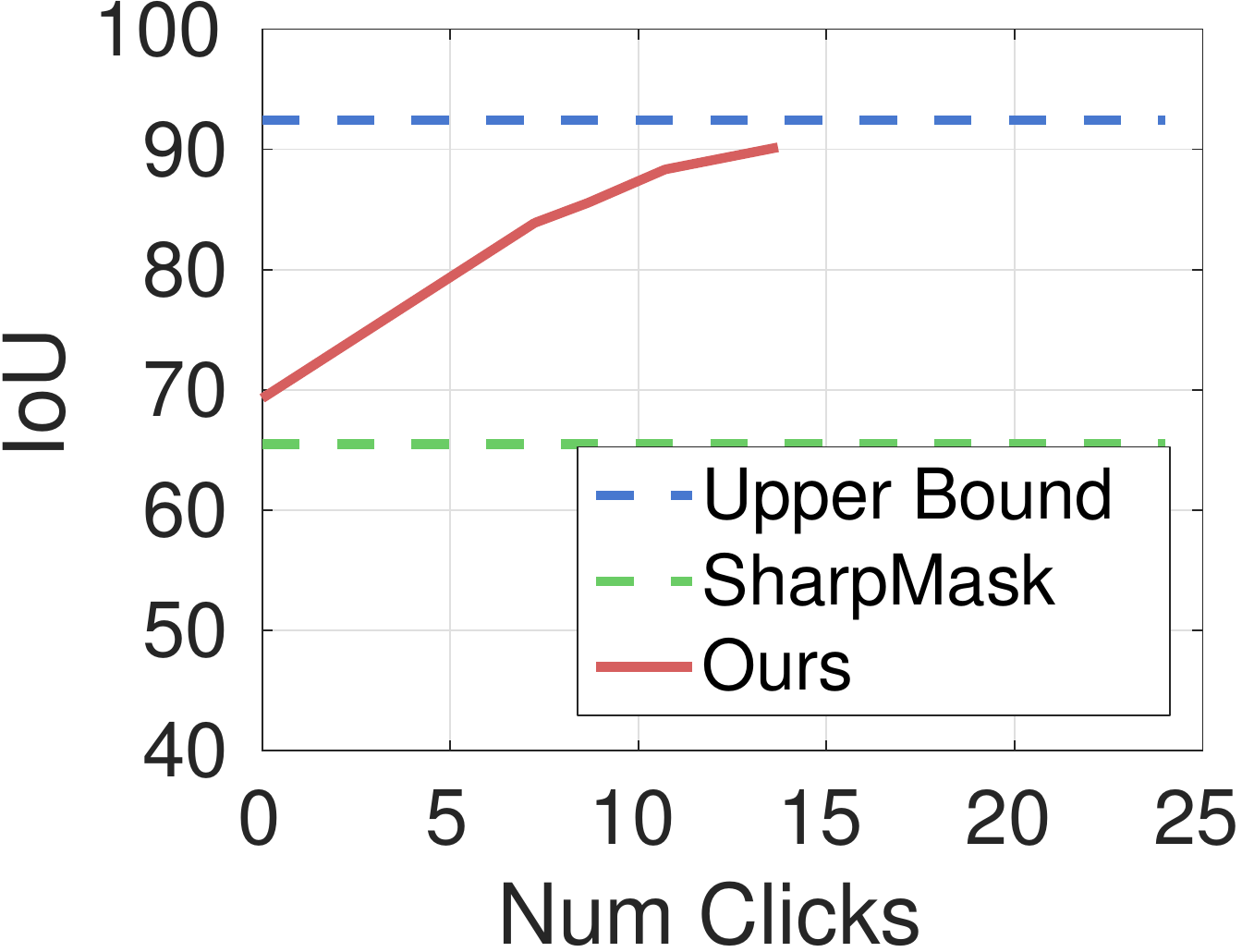} \hspace{-0.5cm} & 
        \includegraphics[height=3.6cm,width=0.25\linewidth,trim=32 0 0 0,clip]{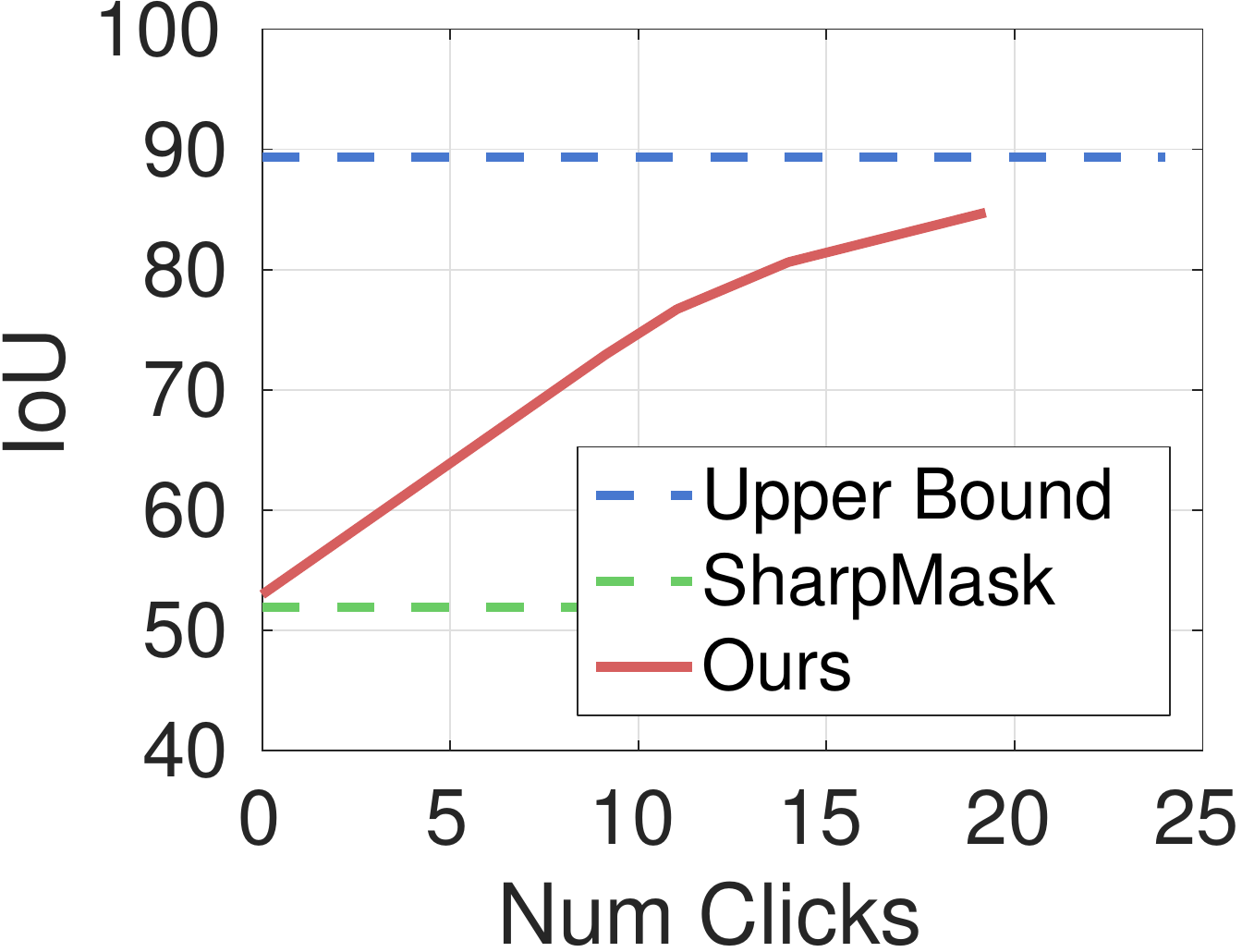} \hspace{-0.5cm} & 
        \includegraphics[height=3.6cm,width=0.25\linewidth,trim=32 0 0 0,clip]{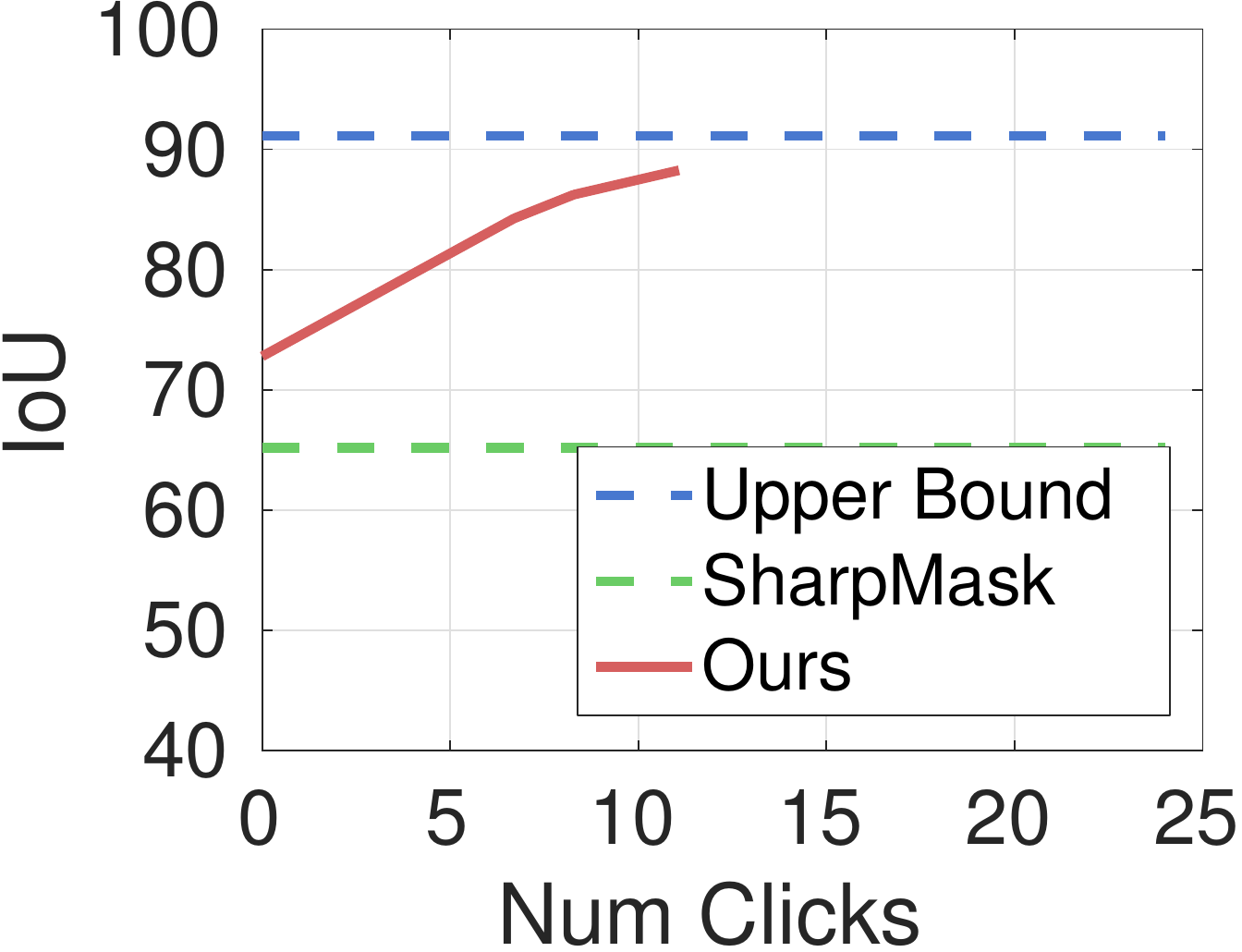} \hspace{-0.5cm} & 
        \includegraphics[height=3.6cm,width=0.25\linewidth,trim=32 0 0 0,clip]{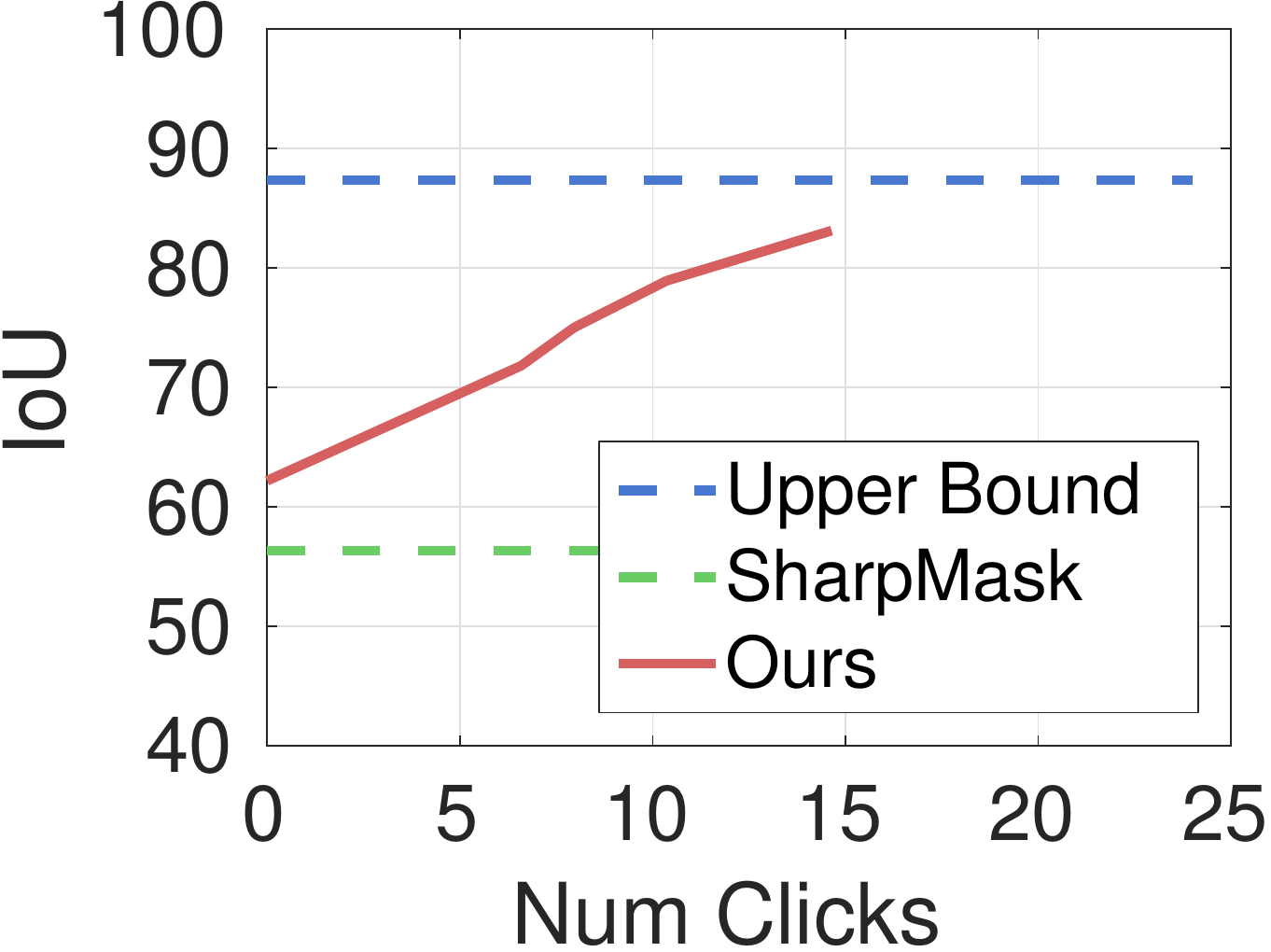} 
    \end{tabular}
    \caption{\textbf{Annotator in the loop:} We show IoU as a function of the number of clicks/corrections.}
    \label{fig:annot_pred_start}
    \vspace{-0.2cm}
\end{figure*}
\end{center}


\begin{table}[t!]
\vspace{-1mm}
\centering
\begin{tabular}{|c|| C{1.9cm} |c|}
	\hline
	Threshold  & Num. Clicks & Mean IOU    \\
	\hline
	1  & 15.79 & 84.74 \\
	2 & 11.77  & 81.43  \\
	3 & 9.39 & 78.40  \\
	4 & 7.86  & 75.79  \\
	\hline
\end{tabular}
\vspace{1.5mm}
\caption{\textbf{Annotator in the loop:} Average number of corrections per instance and IoU, computed across all classes. Threshold indicates chessboard distance to the closest GT vertex.}
\label{table:annotation}
\vspace{-0.0cm}
\end{table}

\begin{table}[t!]
\vspace{-1mm}
\centering
\addtolength{\tabcolsep}{-1.5pt}
\begin{tabular}{|c|| C{1.8cm} |c| C{1.5cm}|}
	\hline
	Method & Num. Clicks & IoU & Annot. Speed-Up  \\
	\hline
	Cityscapes GT & 33.56 & 100 & - \\
	Ann. full image & 79.94 & 69.5 & -  \\
	Ann. crops & 96.09 & 78.6 & - \\
	\hline
	Ours (Automatic) & 0 & 73.3 & No ann. \\
	\hline
	Ours (T=1) & 9.3 & 87.7 & x3.61 \\
	Ours (T=2) & 6.6 & 85.7 & x5.11 \\
	Ours (T=3) & 5.6 & 84.0 & x6.01 \\
	Ours (T=4) & 4.6 & 82.2 & x7.31 \\
	\hline
\end{tabular}
\vspace{1mm}
\caption{\textbf{Our model vs Annotator Agreement}: We hired a highly trained annotator to label \textit{car} instances on additional 10 images (101 instances). We report IoU agreement with Cityscapes GT, and report polygon statistics. We compare our approach with the agreement between the human annotators.}
\label{table:ann-agreement}
\vspace{-0.1cm}
\end{table}

\vspace{-10mm}
To put our results in perspective, we hired an experienced, high-quality annotator. We asked the annotator to annotate all car (including van) instances in 10 randomly selected Cityscapes images from our validation split. We perform two experiments: in the first experiment, the annotator is asked to annotate objects by free-viewing of the full image. In the second experiment, we crop the image patches using the Cityscapes boxes, and place a blue dot on the instance to disambiguate annotation. We take a crop with 15\% of context around the box and scale it to size 224x224. The annotator used the LabelMe tool~\cite{labelme} for annotation. 

In Table ~\ref{table:ann-agreement} we report the IoU achieved by the human annotator as well as the mean number of clicks per instance in each experiment. We can observe that the agreement achieved in IoU is 69.5\% in the free-viewing regime, and 78.60\% when shown the crops (our regime). This number sheds light on what we are to expect from automatic methods in general, and points to some ambiguities in the task. It also indicates that benchmarks should collect multiple annotations of images to reduce the variations and biases across the annotators. We hope our approach will make such data collection feasible and affordable. 

Notice that our model achieves a higher agreement (82\%) by requiring only 4.6 clicks on average, which is a factor of 7.3 speed-up in annotation time. Even at agreement as high as 87.7, the annotation speed-up factor is still 3.6. This showcases the effectiveness of our model as an annotation tool. For all the categories in Cityscapes and following the same procedure, we require only 9.39 clicks on average to obtain 78.40\% IoU agreement, obtaining a speed-up factor of 4.74.

{\bf Comparison with Grabcut}. We also compare the performance of our approach with another semi automatic method on a set of 54 randomly chosen instances. We used the OpenCV implementation of Grabcut~\cite{Rother2004SIGGRAPH} for this experiment. On average, using Grabcut the annotators needed 42.2s and 17.5 clicks per instance, and obtained an average of 70.7\% IoU agreement with the Cityscapes GT. On the same set of images, our model achieves IoUs ranging from 79.7\% to 85.8\%, with 5.0 clicks (T=4) to 9.6 clicks (T=1), respectively. Our expert human annotator needed 87.6 clicks to obtain an IoU of 77.6\% (without using any semi automatic tool). Since our model requires much less human intervention than~\cite{Rother2004SIGGRAPH} (5 vs 17.5 clicks) and requires comparable inference time per click, we expect that in a real world scenario our method would be much faster.

\begin{table}[t!]
\vspace{-1mm}
	\centering
	\begin{tabular}{|c|| c |c|}
		\hline
		Method     &  \# of Clicks & IOU     \\
		\hline
		DeepMask \cite{deepmask}  &  - & 78.3 \\
		SharpMask \cite{sharpmask} &  - & 78.8  \\
		Beat the MTurkers \cite{ChenCVPR14} & 0 & 73.9  \\
		\hline
		Ours (Automatic) & 0 & 74.22 \\
		\hline
		Ours (T=1) & 11.83 & 89.43 \\
		Ours (T=2) & 8.54 & 87.51 \\
		Ours (T=3) & 6.83 & 85.70 \\
		Ours (T=4) & 5.84 & 84.11 \\
		\hline
	\end{tabular}
	\vspace{1mm}
	\caption{Car annotation results on the {\bf KITTI dataset}.}
	\label{table:kitti}
	\vspace{-0mm}
\end{table}

\begin{center}
    \begin{figure*}[h]
    \vspace{-12mm}
    \centering
    \addtolength{\tabcolsep}{-3.6pt}
        \begin{tabular}{c c}
            GT & Ours (Automatic, i.e. 0 corrections) \\
            \includegraphics[width=0.48\linewidth,trim=0 250 0 90,clip]{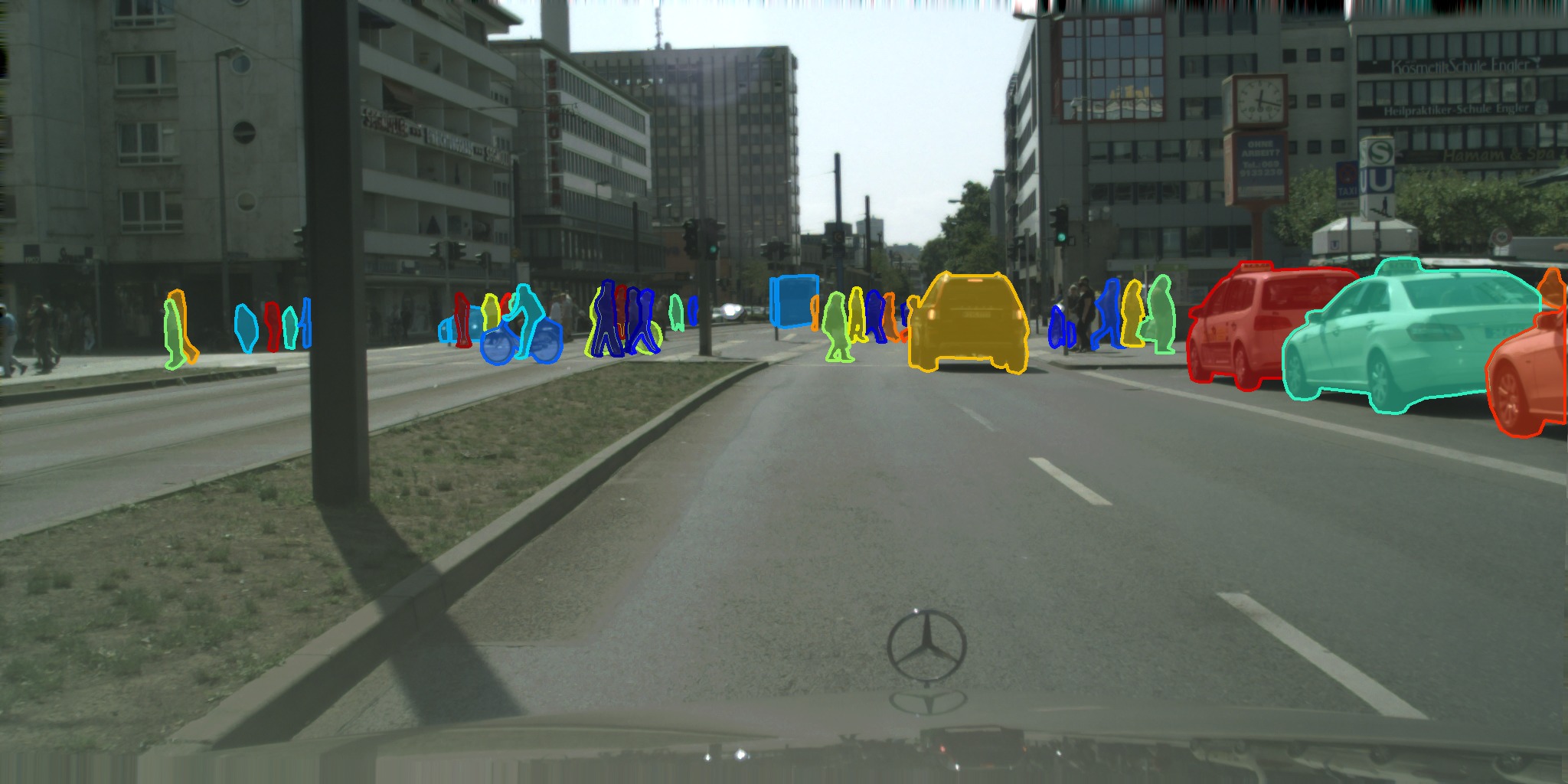} & 
            \includegraphics[width=0.48\linewidth,trim=0 250 0 90,clip]{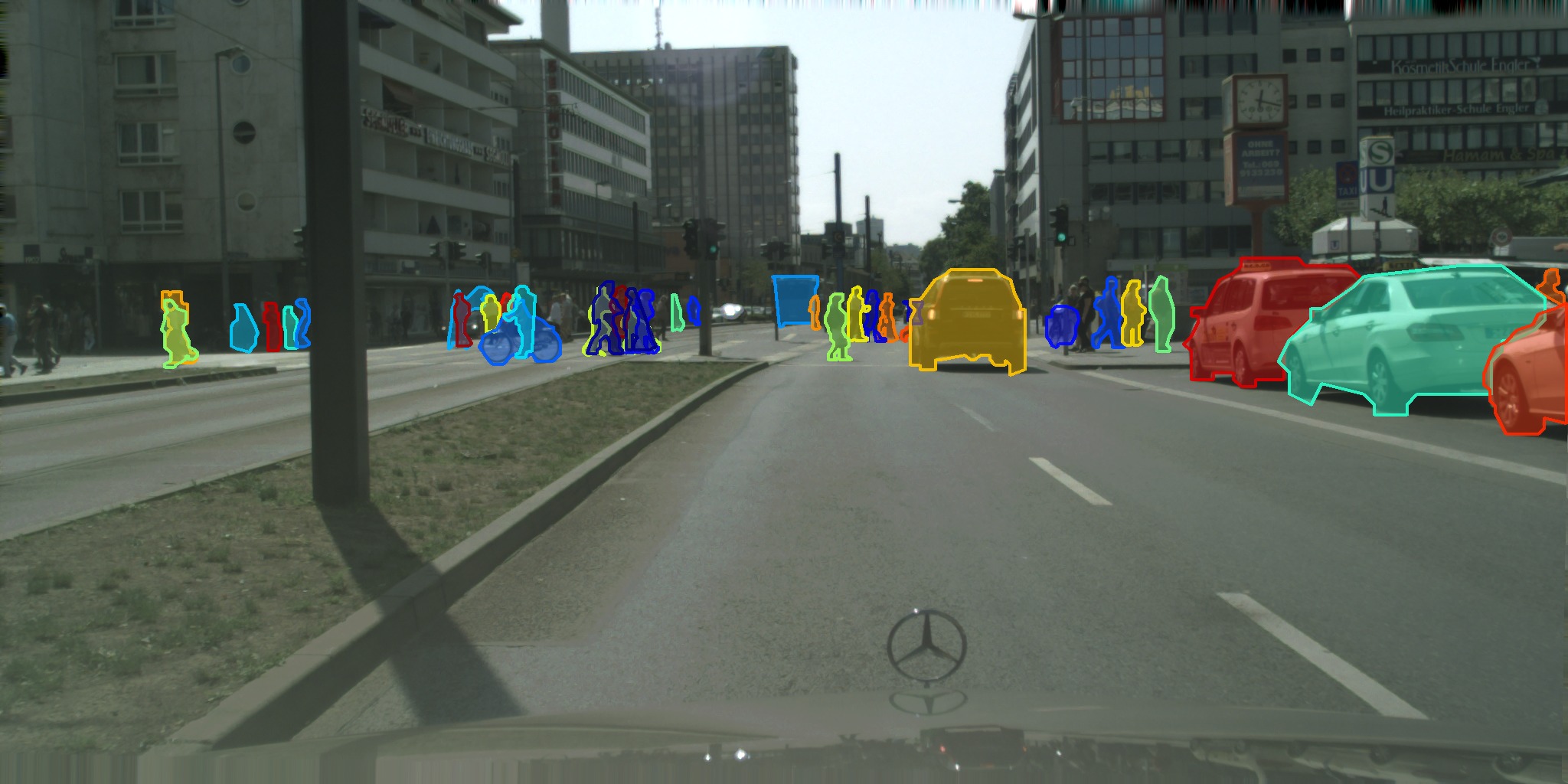}
            \\
            \vspace{-0.6mm}
            \includegraphics[width=0.48\linewidth,trim=0 250 0 90,clip]{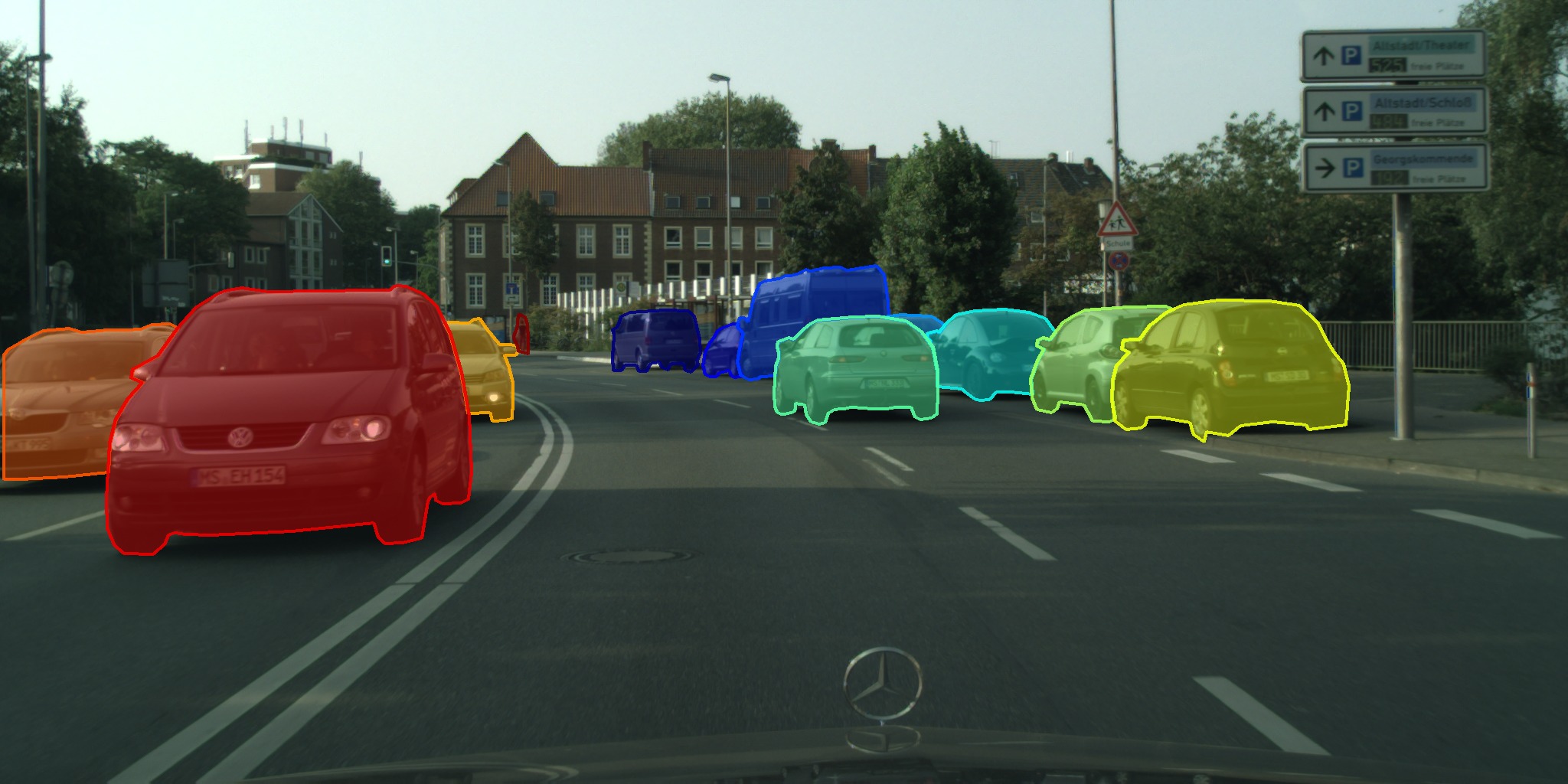} & 
            \includegraphics[width=0.48\linewidth,trim=0 250 0 90,clip]{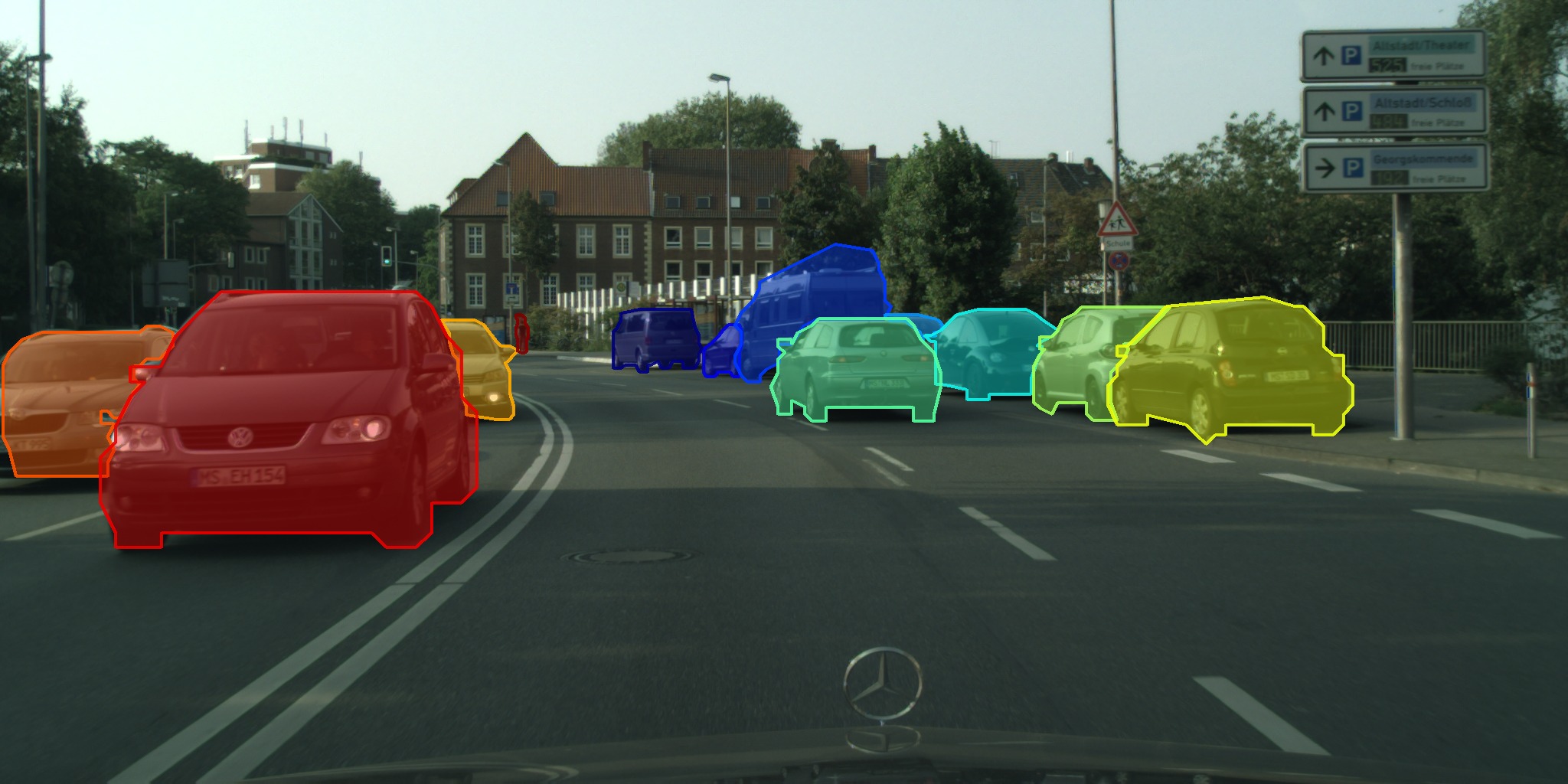} 
            \\
            \vspace{-0.6mm}
            \includegraphics[width=0.48\linewidth,trim=0 250 0 90,clip]{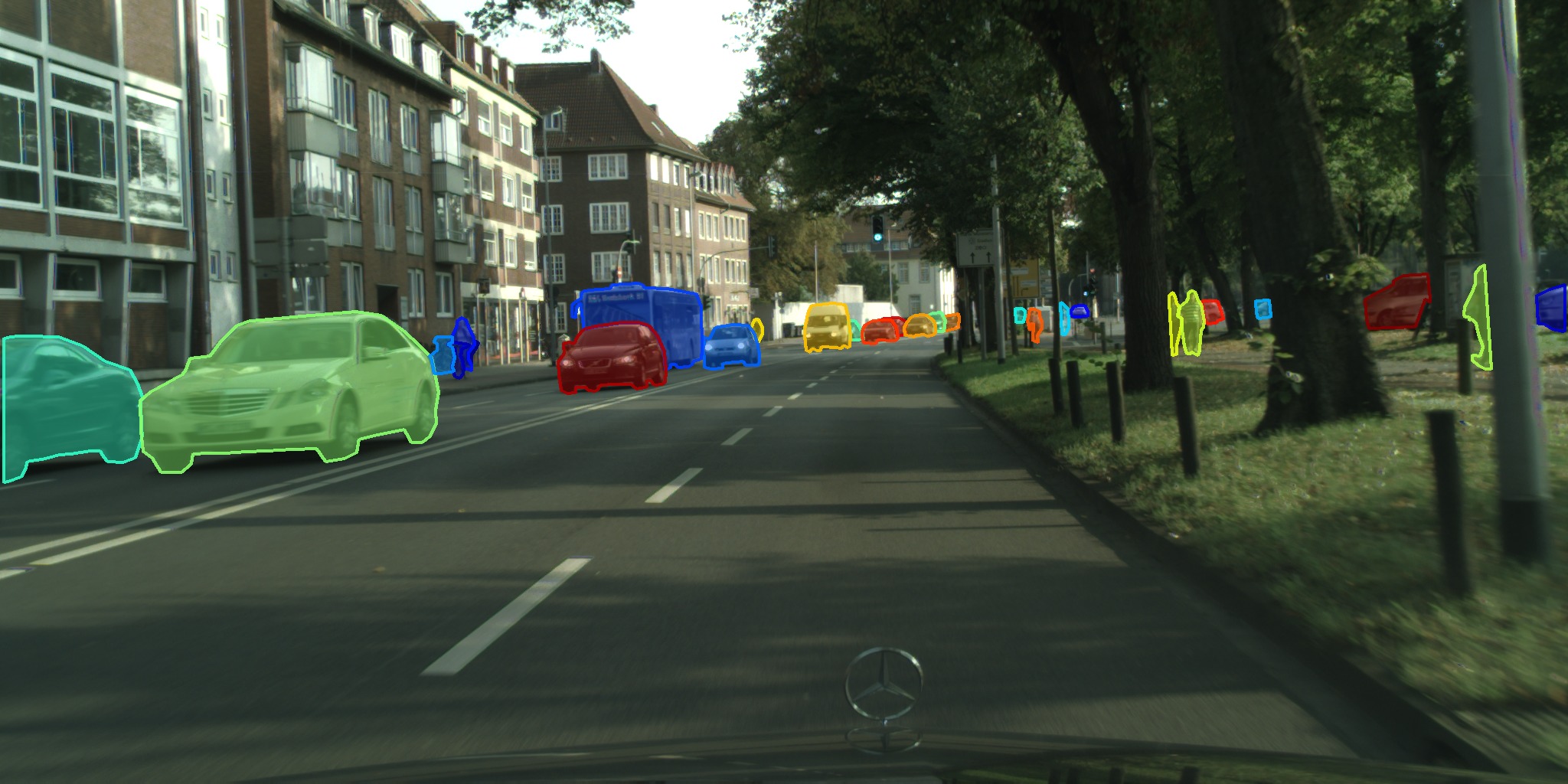} & 
            \includegraphics[width=0.48\linewidth,trim=0 250 0 90,clip]{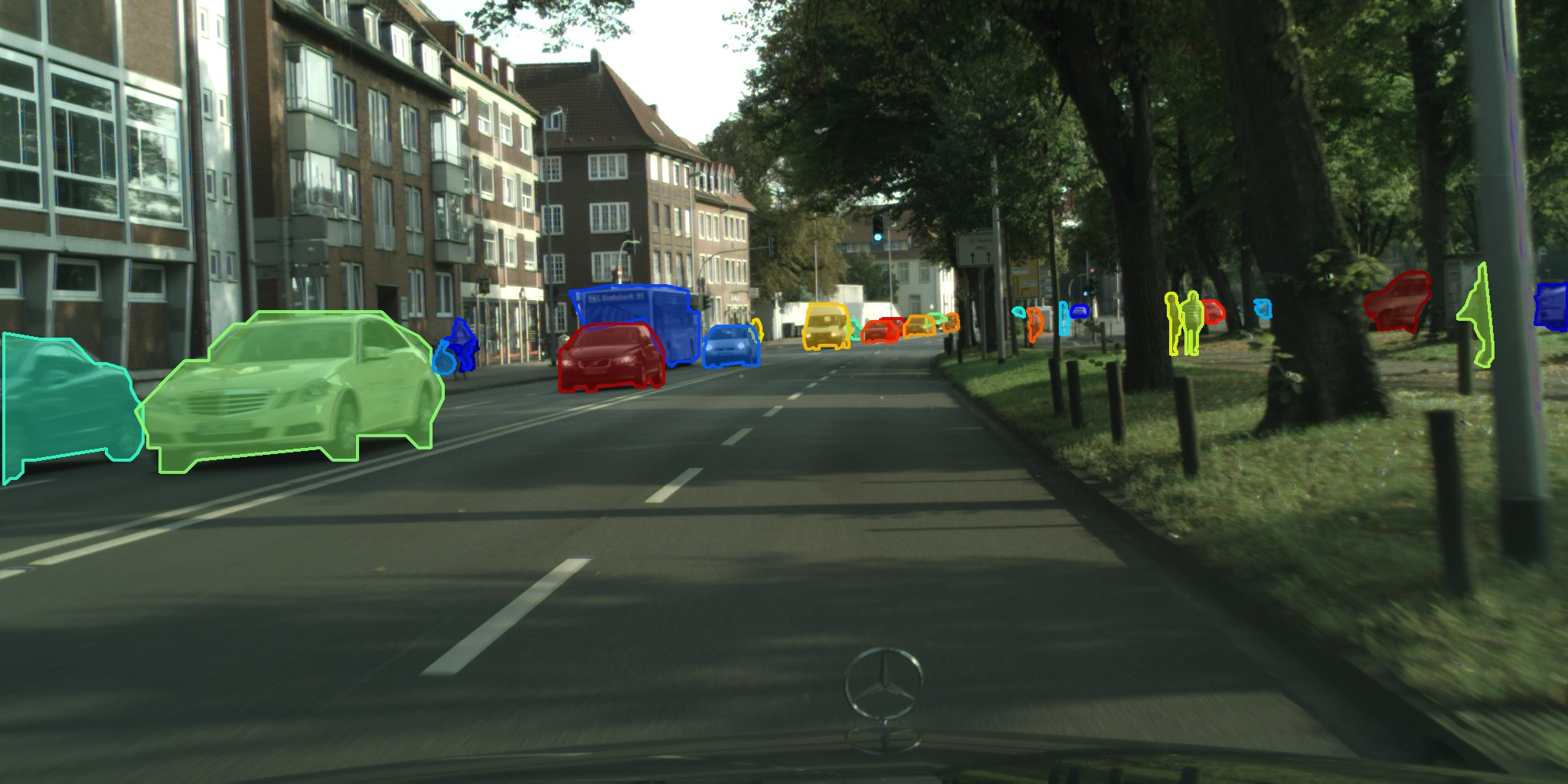} 
            \\
            \vspace{-0.6mm}
            \includegraphics[width=0.48\linewidth,trim=0 250 0 90,clip]{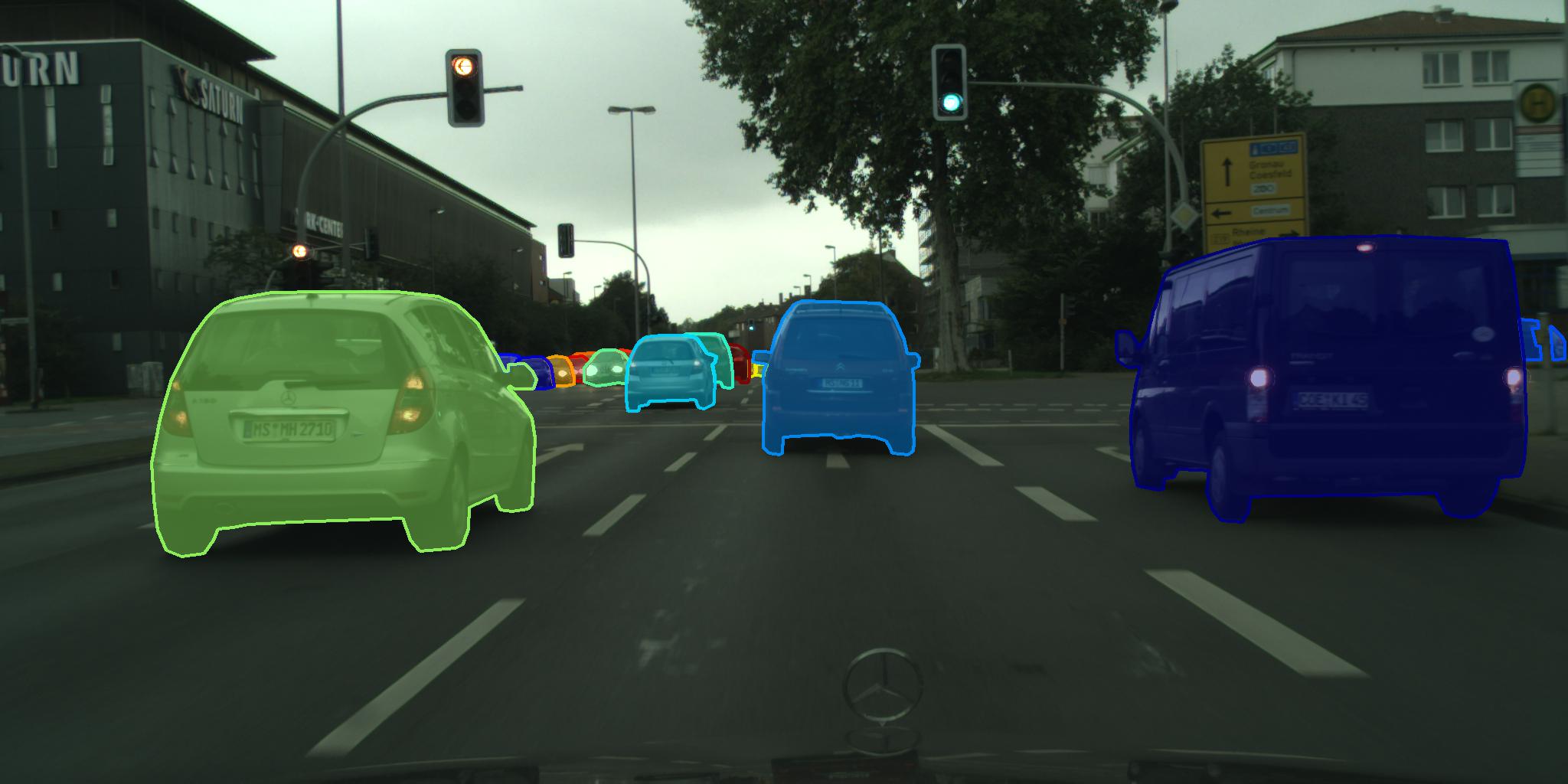} & 
            \includegraphics[width=0.48\linewidth,trim=0 250 0 90,clip]{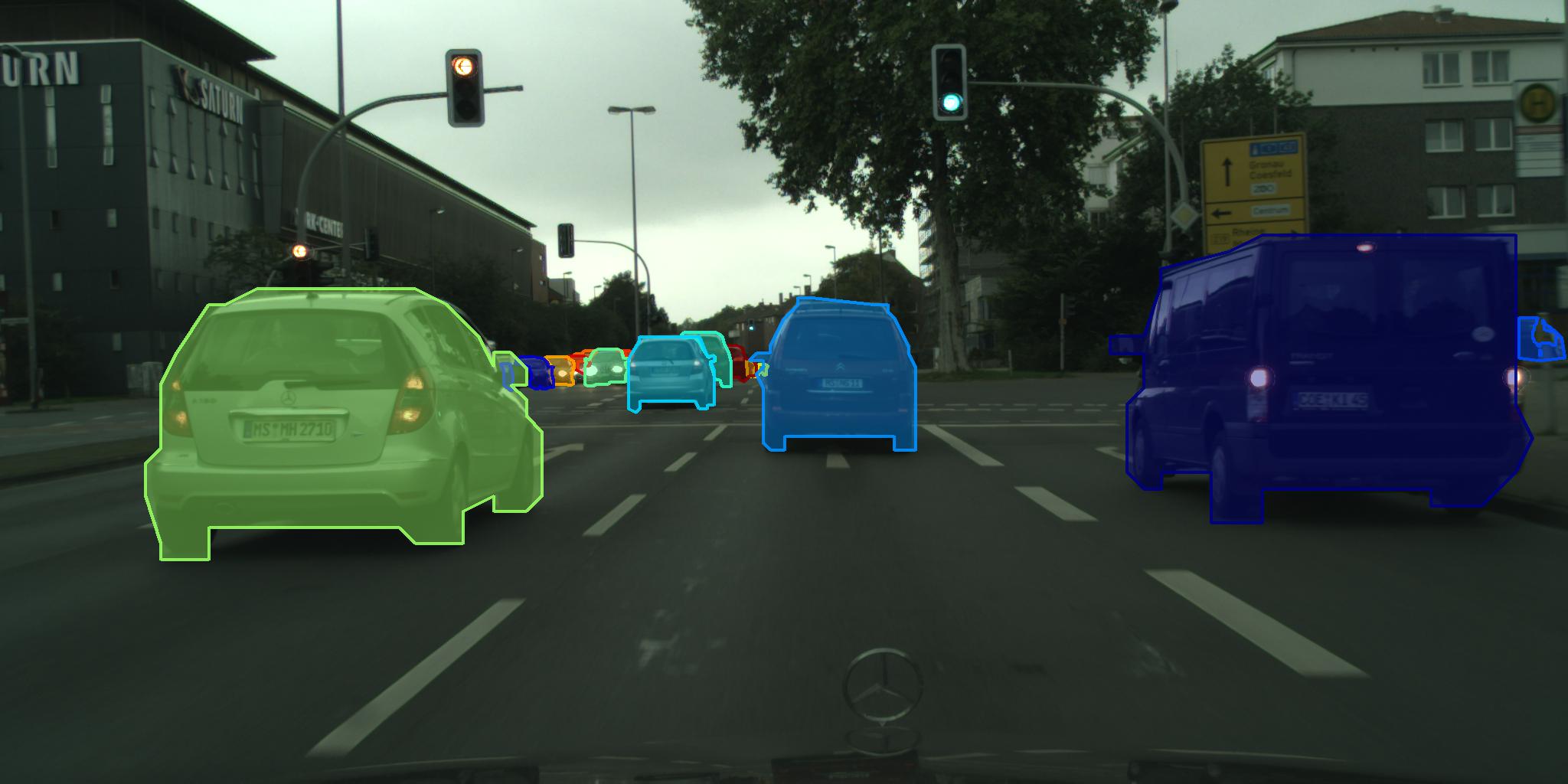} 
        \end{tabular}
        \caption{{\bf Qualitative results in prediction mode}: We show polygons for all classes in the original image. Note that our approach uses GT boxes as input. ({\bf left}) we show the GT labeling of the image, ({\bf right}) we show our polygons without any human intervention. The GT images contain 38, 12, 28 and 16 instances, and required 985, 308, 580 and 338 clicks respectively from their Cityscapes annotators.}
        \label{fig:fullimage}
        \vspace{-1mm}
    \end{figure*}
\end{center}

\vspace{-3.6mm}
\paragraph{Qualitative Results:} In Fig. ~\ref{fig:fullimage} we show examples of images annotated with our method. We remind the reader, that this labeling is obtained by exploiting the GT bounding boxes. In particular, we here show the predictions obtained without any corrections (0 clicks). Our model is able to correctly segment instances with a variety of shapes and sizes. For large instances the quantization error introduced by the output resolution of our model becomes apparent.  Increasing the output resolution is subject of ongoing work. The main challenges are memory considerations as well as challenges due to longer sequences (polygons have more vertices) that the network would need to predict.

\begin{center}
    \begin{figure*}[h]
    \vspace{-1mm}
    \centering
        \begin{tabular}{c c c c} \
        GT & Annotator & Ours (Automatic) & Ours (T=1) \\
            \includegraphics[width=0.162\linewidth]{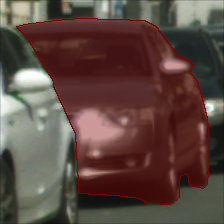} &
            \includegraphics[width=0.162\linewidth]{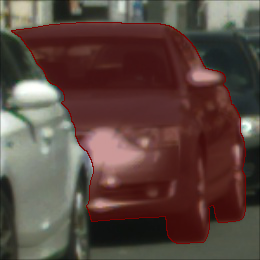} &
            \includegraphics[width=0.162\linewidth]{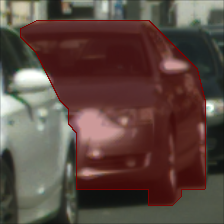} & 
            \includegraphics[width=0.162\linewidth]{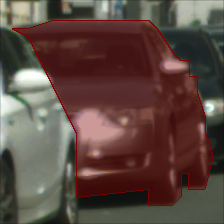} \\
            \vspace{-0.4mm}
            \includegraphics[width=0.162\linewidth]{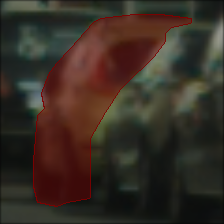} &
            \includegraphics[width=0.162\linewidth]{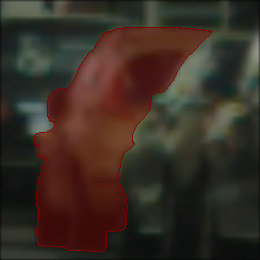} &
            \includegraphics[width=0.162\linewidth]{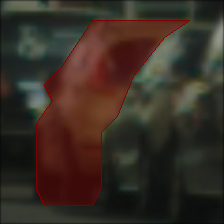} &
            \includegraphics[width=0.162\linewidth]{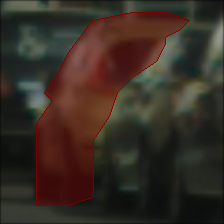} \\
            \vspace{-0.4mm}
            \includegraphics[width=0.162\linewidth]{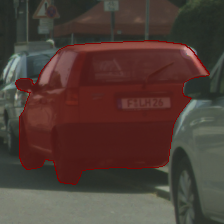} &
            \includegraphics[width=0.162\linewidth]{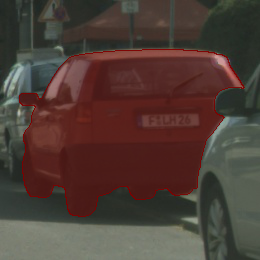} &
            \includegraphics[width=0.162\linewidth]{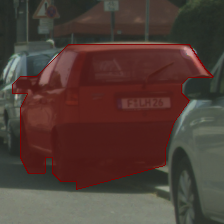} &
	        \includegraphics[width=0.162\linewidth]{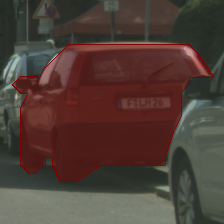} 
        \end{tabular}
        \caption{We look at a few instances in more detail. In the {\bf first column} we show the GT annotation, while in the {\bf second column} we show the polygons from the in-house annotator. We observe that these segmentations are high quality but differ in uncertain areas such as the base of the car. In the {\bf third column} we show the PolygonRNN prediction without human intervention. Finally, in the {\bf fourth column} we show a corrected prediction. We can observe that the segmentation is refined to better outline the car mirrors or wheels.}
        \label{fig:crop}
        \vspace{-8mm}
    \end{figure*}
\end{center}

In Fig.~\ref{fig:crop} we compare annotations of example instances more closely by zooming in on each object. We can inspect the agreement between the GT annotation and our in-house annotator, as well as the quality of the predictions obtained by PolygonRNN with and without corrections. 


\vspace{-18.4mm}
\subsection{Annotating KITTI Instances}
\vspace{-1.5mm}

We also evaluate how well our model that was trained on Cityscapes generalizes to an unseen dataset. We use KITTI for this experiment, which has 741 annotated instances provided by~\cite{ChenCVPR14}. We report the results in Table~\ref{table:kitti}. The object instances in KITTI are usually larger than those found in Cityscapes, making Deepmask and SharpMask perform very similarly. Note that~\cite{ChenCVPR14}, also a method for semi-automatic annotation,  exploited Velodyne point clouds to perform their labeling, which puts it with an unfair advantage. Our model is further penalized by its lower resolution output.  Still, their performance is lower than our fully automatic approach. With only 5.84 clicks on mean per instance our models achieves an IOU comparable to the human annotation agreement, thus reducing the annotation cost.

\vspace{-2mm}
\section{Conclusion}
\label{sec:conc}
\vspace{-1.3mm}

In this paper we proposed an approach to facilitate annotation of object instances. Our Polygon-RNN predicts a polygon outlining an object, and easily incorporates corrections from an annotator in the loop. We show annotation speed-up of factor 4.74 while achieving the same annotation agreement as that between human annotators. The main advantage of our approach is that it produces structurally plausible annotations of objects, and allows us to achieve a desired annotation accuracy by requiring only a few clicks by the annotator. Additional experiments show that our approach generalizes across different datasets, thus showcasing its power as a generic annotation tool.

\vspace{-2mm}
{\small
\section*{Acknowledgement}
\label{sec:ack}
\vspace{-1.3mm}

We acknowledge the support from NSERC, and thank Relu Patrascu for infrastructure support. L.C. was supported by a La Caixa Fellowship.
}
\clearpage
{\small
\bibliographystyle{ieee}
\bibliography{egbib}

\begin{thebibliography}{10}\itemsep=-1pt

\bibitem{ferrari16}
A.~Bearman, O.~Russakovsky, V.~Ferrari, and L.~Fei-Fei.
\newblock What's the point: Semantic segmentation with point supervision.
\newblock {\em arXiv:1506.02106}, 2016.

\bibitem{Boykov2001ICCV}
Y.~Boykov and M.-P. Jolly.
\newblock Interactive graph cuts for optimal boundary \& region segmentation of
  objects in nd images.
\newblock In {\em ICCV}, 2001.

\bibitem{Boykov2004PAMI}
Y.~Boykov and V.~Kolmogorov.
\newblock An experimental comparison of min-cut/max-flow algorithms for energy
  minimization in vision.
\newblock {\em PAMI}, 26(9):1124--1137, 2004.

\bibitem{ChenCVPR14}
L.-C. Chen, S.~Fidler, A.~Yuille, and R.~Urtasun.
\newblock Beat the mturkers: Automatic image labeling from weak 3d supervision.
\newblock In {\em CVPR}, 2014.

\bibitem{chen14semantic}
L.-C. Chen, G.~Papandreou, I.~Kokkinos, K.~Murphy, and A.~L. Yuille.
\newblock Semantic image segmentation with deep convolutional nets and fully
  connected crfs.
\newblock In {\em ICLR}, 2015.

\bibitem{cityscapes}
M.~Cordts, M.~Omran, S.~Ramos, T.~Rehfeld, M.~Enzweiler, R.~Benenson,
  U.~Franke, S.~Roth, and B.~Schiele.
\newblock The cityscapes dataset for semantic urban scene understanding.
\newblock In {\em CVPR}, 2016.

\bibitem{Duan16}
L.~Duan and F.~Lafarge.
\newblock Towards large-scale city reconstruction from satellites.
\newblock In {\em ECCV}, 2016.

\bibitem{pascal-voc-2010}
M.~Everingham, L.~Van~Gool, C.~K.~I. Williams, J.~Winn, and A.~Zisserman.
\newblock The {PASCAL} {V}isual {O}bject {C}lasses {C}hallenge 2010 {(VOC2010)}
  {R}esults.

\bibitem{kitti}
A.~Geiger, P.~Lenz, and R.~Urtasun.
\newblock {Are we ready for Autonomous Driving? The KITTI Vision Benchmark
  Suite}.
\newblock In {\em CVPR}, 2012.

\bibitem{he15deepresidual}
K.~He, X.~Zhang, S.~Ren, and J.~Sun.
\newblock Deep residual learning for image recognition.
\newblock In {\em CVPR}, 2016.

\bibitem{Jain16}
S.~D. Jain and K.~Grauman.
\newblock Active image segmentation propagation.
\newblock In {\em CVPR}, 2016.

\bibitem{adamopt}
D.~Kingma and J.~Ba.
\newblock Adam: A method for stochastic optimization.
\newblock {\em arXiv preprint arXiv:1412.6980}, 2014.

\bibitem{Kuettel2012ECCV}
D.~Kuettel, M.~Guillaumin, and V.~Ferrari.
\newblock Segmentation propagation in imagenet.
\newblock In {\em ECCV}, 2012.

\bibitem{IIS16}
K.~Li, B.~Hariharan, and J.~Malik.
\newblock Iterative instance segmentation.
\newblock In {\em CVPR}, 2016.

\bibitem{Lin16}
D.~Lin, J.~Dai, J.~Jia, K.~He, and J.~Sun.
\newblock Scribblesup: Scribble-supervised convolutional networks for semantic
  segmentation.
\newblock In {\em CVPR}, 2016.

\bibitem{coco}
T.-Y. Lin, M.~Maire, S.~Belongie, J.~Hays, P.~Perona, D.~Ramanan,
  P.~Doll{\'a}r, and C.~L. Zitnick.
\newblock Microsoft coco: Common objects in context.
\newblock In {\em ECCV}, 2014.

\bibitem{LongCVPR2014}
J.~Long, E.~Shelhamer, and T.~Darrell.
\newblock {Fully Convolutional Networks for Semantic Segmentation}.
\newblock arXiv:1411.4038, 2014.

\bibitem{mottaghirole}
R.~Mottaghi, X.~Chen, X.~Liu, N.-G. Cho, S.-W. Lee, S.~Fidler, R.~Urtasun, and
  A.~Yuille.
\newblock The role of context for object detection and semantic segmentation in
  the wild.
\newblock {\em CVPR}, 2014.

\bibitem{Nagaraja}
N.~S. Nagaraja, F.~R. Schmidt, and T.~Brox.
\newblock Video segmentation with just a few strokes.
\newblock In {\em ICCV}, 2015.

\bibitem{deepmask}
P.~O. Pinheiro, R.~Collobert, and P.~Dollar.
\newblock Learning to segment object candidates.
\newblock In {\em NIPS}, pages 1990--1998, 2015.

\bibitem{sharpmask}
P.~O. Pinheiro, T.-Y. Lin, R.~Collobert, and P.~Doll{\'a}r.
\newblock Learning to refine object segments.
\newblock {\em ECCV 2016}, 2016.

\bibitem{Tuset15}
J.~Pont-Tuset, M.~A.~F. Guiu, and A.~Smolic.
\newblock Semi-automatic video object segmentation by advanced manipulation of
  segmentation hierarchies.
\newblock In {\em Intl Workshop on Content-Based Multimedia Indexing}, 2015.

\bibitem{DeepCut}
M.~Rajchl, M.~C. Lee, O.~Oktay, K.~Kamnitsas, J.~Passerat-Palmbach, W.~Bai,
  M.~Damodaram, M.~A. Rutherford, J.~V. Hajnal, B.~Kainz, and D.~Rueckert.
\newblock Deepcut: Object segmentation from bounding box annotations using
  convolutional neural networks.
\newblock In {\em arXiv:1605.07866}, 2016.

\bibitem{torr16}
B.~Romera-Paredes and P.~H.~S. Torr.
\newblock Recurrent instance segmentation.
\newblock In {\em arXiv:1511.08250}, 2015.

\bibitem{Rother2004SIGGRAPH}
C.~Rother, V.~Kolmogorov, and A.~Blake.
\newblock Grabcut: Interactive foreground extraction using iterated graph cuts.
\newblock In {\em SIGGRAPH}, 2004.

\bibitem{labelme}
B.~C. Russell, A.~Torralba, K.~P. Murphy, and W.~T. Freeman.
\newblock Labelme: a database and web-based tool for image annotation.
\newblock {\em International journal of computer vision}, 77(1-3):157--173,
  2008.

\bibitem{vggcnn}
K.~Simonyan and A.~Zisserman.
\newblock Very deep convolutional networks for large-scale image recognition.
\newblock {\em arXiv preprint arXiv:1409.1556}, 2014.

\bibitem{Sun14}
X.~Sun, C.~M. Christoudias, and P.~Fua.
\newblock Free-shape polygonal object localization.
\newblock In {\em ECCV}, 2014.

\bibitem{Uhrig16}
J.~Uhrig, M.~Cordts, U.~Franke, and T.~Brox.
\newblock Pixel-level encoding and depth layering for instance-level semantic
  labeling.
\newblock In {\em arXiv:1604.05096}, 2016.

\bibitem{convlstm}
S.~Xingjian, Z.~Chen, H.~Wang, D.-Y. Yeung, W.-k. Wong, and W.-c. Woo.
\newblock Convolutional lstm network: A machine learning approach for
  precipitation nowcasting.
\newblock In {\em NIPS}, pages 802--810, 2015.

\bibitem{JiaXu14}
J.~Xu, A.~Schwing, and R.~Urtasun.
\newblock Tell me what you see and i will show you where it is.
\newblock In {\em CVPR}, 2014.

\bibitem{Yamaguchi12}
K.~Yamaguchi, M.~H. Kiapour, L.~E. Ortiz, and T.~L. Berg.
\newblock Parsing clothing in fashion photographs.
\newblock In {\em CVPR}, 2012.

\bibitem{dilation}
F.~Yu and V.~Koltun.
\newblock Multi-scale context aggregation by dilated convolutions.
\newblock {\em arXiv preprint arXiv:1511.07122}, 2015.

\bibitem{ZhangCVPR16}
Z.~Zhang, S.~Fidler, and R.~Urtasun.
\newblock Instance-level segmentation for autonomous driving with deep densely
  connected mrfs.
\newblock In {\em CVPR}, 2016.

\bibitem{ZhangCVPR12}
Z.~Zhang, S.~Fidler, J.~W. Waggoner, Y.~Cao, J.~M. Siskind, S.~Dickinson, and
  S.~Wang.
\newblock Super-edge grouping for object localization by combining appearance
  and shape information.
\newblock In {\em CVPR}, 2012.

\bibitem{ZhangICCV15}
Z.~Zhang, A.~Schwing, S.~Fidler, and R.~Urtasun.
\newblock Monocular object instance segmentation and depth ordering with cnns.
\newblock In {\em ICCV}, 2015.

\bibitem{ade20k}
B.~Zhou, H.~Zhao, X.~Puig, S.~Fidler, A.~Barriuso, and A.~Torralba.
\newblock Semantic understanding of scenes through ade20k dataset.
\newblock In {\em arXiv:1608.05442}, 2016.

\end{thebibliography}
}

\end{document}